\documentclass[11pt,a4paper]{extarticle}
\usepackage{styles/technical_report}
\usepackage{amsmath,amssymb,booktabs}
\usepackage{caption}
\usepackage{graphicx}
\usepackage{subcaption}
\definecolor{capLightGray}{HTML}{888888}
\definecolor{capMediumGray}{HTML}{676767}
\definecolor{capDarkGray}{HTML}{475467}
\definecolor{capPlotBlue}{HTML}{4679DE}
\definecolor{capTokenGreen}{HTML}{087A55}
\definecolor{capTokenRed}{HTML}{B42318}
\definecolor{capCaseGreen}{HTML}{2E7D50}
\definecolor{capCaseRed}{HTML}{B83A2D}
\usepackage{algorithm}
\usepackage{algpseudocode}
\usepackage{multirow}
\usepackage{listings}
\definecolor{algcomment}{RGB}{20,77,89}
\definecolor{algkeyword}{RGB}{35,55,125}
\definecolor{algfunction}{RGB}{82,51,110}
\lstdefinestyle{paperpython}{
  language=Python,
  basicstyle=\small\ttfamily,
  numbers=none,
  frame=none,
  linewidth=\linewidth,
  columns=fullflexible,
  keepspaces=true,
  showstringspaces=false,
  showspaces=false,
  showtabs=false,
  breaklines=false,
  tabsize=4,
  aboveskip=5pt,
  belowskip=4pt,
  keywordstyle=\color{algkeyword},
  commentstyle=\color{algcomment},
  stringstyle=\color{black!75},
  emph={rollout,judge,clone,actor_snapshot,
    count_calls,fulltrace_pair,valid_pair,
    no_grad,token_logp,valid_logp,mean,clip,
    exp,relu,last_generated,group_normalize,
    grpo_update,generated_mask},
  emphstyle=\color{algfunction}
}
\usepackage{tikz}
\usepackage{longtable}
\usepackage{colortbl}
\usetikzlibrary{arrows.meta,shapes.geometric,positioning,calc}
\definecolor{paperBlue}{HTML}{4285F4}
\definecolor{paperBlueLight}{HTML}{EEF5FF}
\definecolor{paperMint}{HTML}{4FCFC1}
\definecolor{paperLavender}{HTML}{6C63FF}
\definecolor{paperOrange}{HTML}{E8A05B}
\definecolor{paperPanel}{HTML}{F3F4F5}
\definecolor{paperGrid}{HTML}{E4E7EB}
\definecolor{paperInk}{HTML}{354052}
\usepackage{amsthm}
\newtheorem{lemma}{Lemma}
\newtheorem{proposition}{Proposition}
\newcommand{\papertitle}{\method{}: Tool-Attributed Regularization for Search-Efficient Multimodal Agents}
\DeclareRobustCommand{\method}{\texorpdfstring{\textrm{\textbf{\textsc{TAR}}}}{TAR}}
\usepackage{flafter}
\usepackage{placeins}

\usepackage{hyperref}
\definecolor{reportReferencePink}{HTML}{FF6B7C}
\hypersetup{colorlinks=true,linkcolor=reportElectricBlue,citecolor=reportReferencePink,
  urlcolor=reportReferencePink,filecolor=reportReferencePink,pdfborder={0 0 0},
  pdftitle={TAR: Tool-Attributed Regularization for Search-Efficient Multimodal Agents},
  pdfauthor={Yang Xu, Shaohang Wei, Yongcan Yu, Houfeng Wang, Kai Ming Ting, Asher Qin, Luke Mao, Lei Li}}
\newcommand{\ReportDate}{2026-11-30}
\newcommand{\ProfileAuthor}[3]{\ReportAuthor{\href{https://openreview.net/profile?id=\string~#2}{\textcolor{black}{#1}}}{#3}}
\title{\method{}: Tool-Attributed Regularization for\\Search-Efficient Multimodal Agents}
\author{%
  \ProfileAuthor{Yang Xu}{Yang_Xu26}{1,*}, \ProfileAuthor{Shaohang Wei}{Shaohang_Wei1}{2,*},
  \ProfileAuthor{Yongcan Yu}{Yongcan_Yu1}{3}, \ProfileAuthor{Houfeng Wang}{Houfeng_Wang1}{2},\\
  \ProfileAuthor{Kai Ming Ting}{Kai_Ming_Ting2}{1}, \ProfileAuthor{Asher Qin}{Asher_Qin1}{4},
  \ProfileAuthor{Luke Mao}{lukemao1}{4}, \ProfileAuthor{Lei Li}{Lei_Li23}{4}}
\newcommand{\ReportAffiliations}{%
  \mbox{\textsuperscript{1}Nanjing University}\hspace{1.8em}%
  \mbox{\textsuperscript{2}Peking University}\hspace{1.8em}%
  \mbox{\textsuperscript{3}CASIA}\hspace{1.8em}%
  \mbox{\textsuperscript{4}Tencent}}
\begin{document}
\maketitle
\let\ReportDefaultFootnoteRule\footnoterule
\renewcommand{\footnoterule}{\kern-3pt\hrule width12pc height0.4pt\kern2.6pt}
\begingroup
\renewcommand{\thefootnote}{\fnsymbol{footnote}}
\footnotetext[1]{Equal contribution.}
\endgroup
\addtocontents{toc}{\protect\setcounter{tocdepth}{-1}}
\begin{minipage}{\textwidth}
\centering
\captionsetup{type=figure,hypcap=false}
\includegraphics[width=\textwidth]{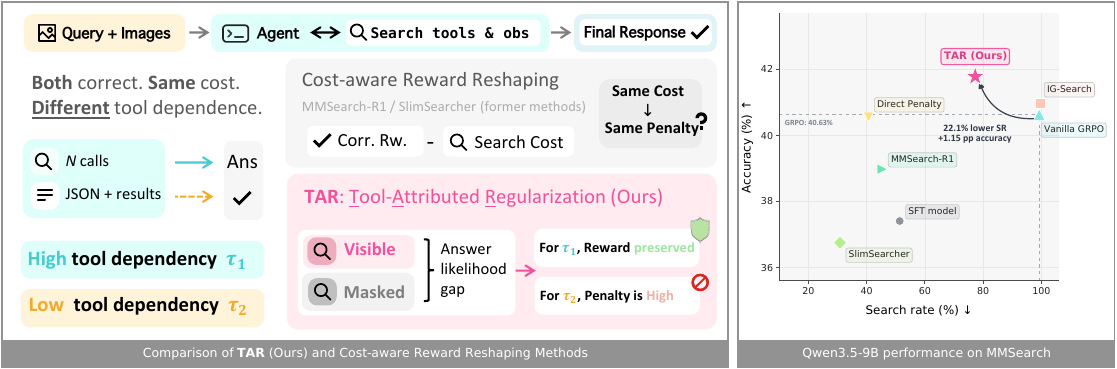}
\caption{%
\textbf{\method{} overview: allocating search penalties by response support rather than cost alone.}
Left: for correct trajectories with the same search cost, cost-aware reward reshaping applies identical penalties, whereas \method{} compares response likelihoods under visible and masked search traces to preserve reward under high tool dependency ($\tau_1$) and penalize low dependency ($\tau_2$).
Right: accuracy and search rate on MMSearch with Qwen3.5-9B (Table~\ref{tab:main}), where \method{} reduces search rate from $99.26\%$ to $77.29\%$ while improving accuracy from $40.63\%$ to $41.78\%$ over Vanilla GRPO.%
}
\label{fig:headline}
\end{minipage}\par

\begin{abstract}
External search helps multimodal agents answer questions beyond their parametric knowledge, yet excessive retrieval inflates tool-use costs.
Penalizing search costs during training reduces call frequency, but treats equally costly correct trajectories alike, discouraging useful retrieval alongside redundant calls.
We introduce \textbf{T}ool-\textbf{A}ttributed \textbf{R}egularization (\method{}), a training-time method that assigns search penalties according to how strongly the interaction trace supports the final response.
Keeping the original multimodal input fixed, \method{} uses the agent itself to compare the likelihood of a completed response with and without its intermediate trace, yielding a bounded Tool Credit Score.
During reinforcement learning, \method{} penalizes only correct searches with weak trace support and adds no extra cost at inference time.
Across SimpleVQA, DynVQA, and MMSearch on Qwen3.5-9B and Qwen3.5-35B-A3B, \method{} lowers search rate by nearly $20\%$ (from over $95\%$ to $76\%$--$79\%$) relative to Vanilla GRPO while keeping average accuracy within $0.5\%$.
\method{} also outperforms existing cost-penalized baselines by $1.94\%$--$13.60\%$ in average accuracy across both backbones.
\end{abstract}

\section{Introduction}
\label{sec:intro}
Multimodal agents built on large vision-language models (LVLMs) use image and web search to acquire up-to-date information beyond their parametric knowledge~\citep{hu2023avis,wu2025mmsearchr1,narayan2025deepmmsearch}.
Retrieved evidence enables these agents to identify unfamiliar visual entities and answer knowledge-intensive questions~\citep{hu2023avis,narayan2025deepmmsearch}.
With reinforcement learning (RL), agents can learn search strategies directly through interaction, so how rewards are designed shapes when they invoke external tools~\citep{jin2025searchr1,chen2025research,wu2025mmsearchr1}.

However, outcome-only methods such as Search-R1~\citep{jin2025searchr1}, ReSearch~\citep{chen2025research}, and DeepMMSearch-R1~\citep{narayan2025deepmmsearch} reward answer correctness without penalizing search cost, leading agents to search on nearly every query and inflating interaction overhead.
To curb excessive search, cost-aware methods penalize tool use directly: MMSearch-R1~\citep{wu2025mmsearchr1} discounts any correct answer that uses search, SlimSearcher~\citep{slim2026} scales rewards by relative cost and length among successful rollouts, and HyperEyes~\citep{li2026hypereyes} penalizes deviations from reference costs.
Yet two correct trajectories with the same search cost can rely on retrieval very differently: one may need web or visual search to resolve an unfamiliar entity or recent event, whereas another answers a question already solvable from the input image alone (Figure~\ref{fig:headline}, left).
Penalizing both equally discourages useful retrieval.
Under Group Relative Policy Optimization (GRPO)~\citep{shao2024deepseekmath}, simply lowering a cost-only penalty weight still fails to separate helpful from redundant searches within a rollout group, motivating penalty assignment by how much the search supports the answer rather than by cost alone.

We propose Tool-Attributed Regularization (\method{}), a training-time method that assigns search penalties according to how strongly the interaction trace supports the completed response.
Unlike step-level rewards based on gold-answer likelihoods~\citep{igsearch2026,tips2026,zhu2026lotapo}, \method{} draws on context attribution and self-likelihood comparison~\citep{cohenwang2024contextcite,han2025self} to evaluate the agent's own completed response directly.
For each correct search rollout, the agent scores its final response tokens with and without the intermediate trace under the same multimodal input, mapping the length-normalized log-likelihood drop to a bounded Tool Credit Score.
A large likelihood drop when removing the trace indicates strong support and leaves the full reward intact, whereas a small change marks a weakly supported search and triggers a penalty during GRPO training, with no extra scoring at inference time.
Removing the intermediate trace as a whole provides a clean trajectory-level support signal that separates helpful from redundant searches of equal cost under group normalization.

We evaluate \method{} on Qwen3.5-9B and Qwen3.5-35B-A3B~\citep{qwen2026base9b,qwen2026base35b} across 4,236 examples from SimpleVQA~\citep{cheng2025simplevqa}, DynVQA~\citep{liu2024dynvqa}, and MMSearch~\citep{jiang2024mmsearch} (including MMSearch-Plus~\citep{tao2025mmsearchplus}).
Relative to Vanilla GRPO, \method{} reduces aggregate search rate by $19.74\%$ and $19.62\%$ (from $98.70\%$ and $95.54\%$ to $78.97\%$ and $75.92\%$) while changing average accuracy by only $-0.05\%$ and $+0.40\%$ on the two backbones (Figure~\ref{fig:headline}, right).
In contrast, IG-Search~\citep{igsearch2026} matches Vanilla GRPO in accuracy only by searching on over $95\%$ of questions, whereas \method{} outperforms existing cost-penalized baselines by $1.94\%$--$13.60\%$ in average accuracy.
Comparisons on skipped-search questions and across penalty weights further show that \method{} stays within $0.5\%$ of Vanilla GRPO on the roughly $20\%$ of queries where it skips search, whereas cost-only penalties lose accuracy even at small weights.

\begin{ContributionBox}
\textbf{Our main contributions are:}
\begin{itemize}
\item We propose \method{}, a training-time regularization method that compares the likelihood of an agent's own response with and without its interaction trace to penalize only weakly supported correct searches, with no inference overhead.
\item We analyze penalty assignment under GRPO, showing when group normalization cancels uniform cost scaling in equal-reward groups and how support-based penalties separate equally costly successful searches.
\item Across two Qwen3.5 backbones and three multimodal search benchmarks, \method{} lowers search rate by nearly $20\%$ while staying within $0.5\%$ of Vanilla GRPO in average accuracy, preserving accuracy when skipping search and across penalty weights.
\end{itemize}
\end{ContributionBox}

\let\footnoterule\ReportDefaultFootnoteRule
\ReportNeedspace{75pt}
\section{Related Work}
\label{sec:related}
\paragraph{Multimodal search agents.}
Retrieval-augmented generation supplies external evidence for answering~\citep{lewis2020retrieval,lin2022ravqa}, while interactive agents acquire and use evidence through tool calls~\citep{nakano2021webgpt,yao2022react}.
Search-R1~\citep{jin2025searchr1} and ReSearch~\citep{chen2025research} train search policies with reinforcement learning.
For visual information seeking, AVIS~\citep{hu2023avis} guides tool selection through a planner, while MMSearch-R1~\citep{wu2025mmsearchr1} and DeepMMSearch-R1~\citep{narayan2025deepmmsearch} learn image and text search policies from outcome rewards.
On-policy distillation also uses visual agents' own trajectories to diagnose errors in acquiring or interpreting tool observations~\citep{wei2026visualevidence}.

\paragraph{Adaptive retrieval and search cost.}
Retrieval decisions can depend on generation confidence, learned reflection, question complexity, or current information needs~\citep{jiang2023flare,asai2023selfrag,jeong2024adaptiverag,su2024dragin}.
Cost-aware objectives incorporate explicit search penalties: MMSearch-R1~\citep{wu2025mmsearchr1} discounts the accuracy reward whenever a correct answer uses search, SlimSearcher~\citep{slim2026} scales rewards by relative tool cost and generated length among successful trajectories, and HyperEyes~\citep{li2026hypereyes} uses question-specific reference costs.
ProMMSearchAgent~\citep{yan2026promm} uses knowledge-boundary annotations to reward search decisions.
\method{} instead assigns penalties by how strongly the trace supports the agent's completed response, allowing equally costly correct trajectories to receive different training rewards depending on whether retrieval actually helped.

\paragraph{Likelihood feedback and context attribution.}
Toolformer~\citep{toolformer2023} selects tool-use examples through prediction-loss improvement.
IG-Search~\citep{igsearch2026} compares gold-answer likelihood under retrieved and random documents for query-level credit, whereas TIPS~\citep{tips2026} uses increases in a valid-answer likelihood potential across turns~\citep{ng1999shaping}.
LOTAPO~\citep{zhu2026lotapo} replaces one search turn and its observation with a deletion placeholder while preserving downstream interactions to score gold-answer likelihood.
TAPO~\citep{tapo2026} transfers credit from successful rollouts to parameter-matched tool steps in failed ones.
SAR~\citep{han2025self} adds a query-conditioned relative perplexity reduction to verifiable rewards to encourage concise reasoning.
Context attribution and rationale evaluation likewise study how supplied evidence affects model outputs by removing or masking context spans~\citep{cohenwang2024contextcite,deyoung2020eraser}.
Building on these ideas, \method{} adopts SAR's relative-perplexity form while retaining the multimodal input in both branches and removing the intermediate trace to score the agent's own response.
Rather than using turn-level process rewards, \method{} applies this trajectory-level support score during training to penalize only correct searches with weak support, adding no inference cost.

\section{\method{}: Tool-Attributed Regularization}
\label{sec:method}
\begin{figure}[!t]
\centering
\includegraphics[width=\textwidth]{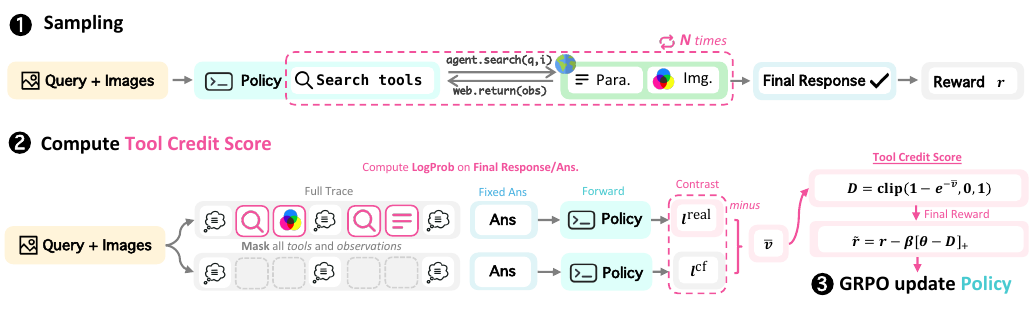}
\caption{%
\textbf{Overview of the \method{} training pipeline.}
\textbf{(1) Sampling:} Given a multimodal input (query and images), the policy interacts with web and vision search tools over up to $N$ turns to generate a final response and receive a correctness reward $r$.
\textbf{(2) Compute Tool Credit Score:} For eligible correct, tool-using trajectories, a detached policy snapshot computes teacher-forced mean token log-likelihoods on the same fixed final response under the full interaction trace ($\ell^{\mathrm{real}}$) and the input-only context with the intermediate trace removed ($\ell^{\mathrm{cf}}$), retaining the original images in both branches.
Their contrast $\bar v = \ell^{\mathrm{real}} - \ell^{\mathrm{cf}}$ defines the bounded Tool Credit Score $D = \operatorname{clip}(1 - e^{-\bar v}, 0, 1)$.
\textbf{(3) GRPO Policy Update:} Trajectories with $D < \theta$ incur a terminal penalty, yielding the regularized reward $\widetilde r = r - \beta[\theta - D]_+$ for group-relative policy optimization (GRPO), while paired scoring is omitted at inference time.%
}
\label{fig:framework}
\end{figure}

Outcome-only rewards encourage unnecessary calls, whereas cost-only penalties discourage helpful retrieval.
During GRPO training~\citep{shao2024deepseekmath}, \method{} scores how strongly each trace supports the completed response and penalizes only low-support correct searches (Figure~\ref{fig:framework}), with zero inference overhead.

\ReportNeedspace{113pt}
\subsection{Multimodal Search Rollout and Problem Setup}
\label{sec:formulation}\label{sec:motivation}
Given a multimodal input $x=(I,q)$ with serialized instructions, the policy $\pi_\phi$ answers directly or invokes Web and Vision search across turns, appending textual observations to its context.
For each input, we sample a group $\mathcal G=\{\tau_i\}_{i=1}^m$ and obtain correctness rewards $r_i\in\{0,1\}$.
Since $r_i$ and call counts do not show how much the answer relied on search, we write $\tau_i=(h_i,a_i)$ to separate the intermediate trace $h_i$ (reasoning, tool calls, and observations) from the final response segment $a_i$ generated in the answering turn (extraction details in Appendix~\ref{sec:dependence-implementation}).

\subsection{Tool Credit Score}
\label{sec:paired-scoring}
In open-ended QA, reference-answer scoring methods such as IG-Search~\citep{igsearch2026} and LOTAPO~\citep{zhu2026lotapo} suffer from surface-form mismatch, while regenerating counterfactual rollouts is costly and noisy.
We therefore fix the actor's own completed response $a_i$ and construct two contexts:
\begin{equation}
C_i^{\mathrm{real}}=x\oplus h_i,\qquad C_i^{\mathrm{cf}}=x.
\label{eq:contexts}
\end{equation}
Both branches retain $x$ and evaluate identical tokens $a_i$ under a detached actor snapshot $\pi_{\bar\phi}$, with $C_i^{\mathrm{cf}}$ removing $h_i$ and rebuilding positions without regeneration.
To avoid confounding support with varying lengths $L_i=|a_i|$, we compute the length-normalized log-likelihood:
\begin{equation}
\ell_i^b=\frac{1}{L_i}\sum_{j=1}^{L_i}\log\pi_{\bar\phi}(a_{i,j}\mid C_i^b,a_{i,<j}),\quad b\in\{\mathrm{real},\mathrm{cf}\}.
\label{eq:paired-logp}
\end{equation}
Let $v_{i,j}=\log\pi_{\bar\phi}(a_{i,j}\mid C_i^{\mathrm{real}},a_{i,<j})-\log\pi_{\bar\phi}(a_{i,j}\mid C_i^{\mathrm{cf}},a_{i,<j})$ denote the token-level log-probability change between the full-trace and trace-masked forward passes.
Since the mean difference $\bar v_i$ is unbounded above, we map it to a bounded Tool Credit Score (TCS) $D_i$:
\begin{equation}
\bar v_i=\frac{1}{L_i}\sum_{j=1}^{L_i}v_{i,j}=\ell_i^{\mathrm{real}}-\ell_i^{\mathrm{cf}},\quad
D_i=\operatorname{clip}(1-e^{-\bar v_i},0,1).
\label{eq:d}
\end{equation}
This transform compares geometric-mean token likelihoods and maps $\bar v_i\le 0$ to $0$.
Because both passes score identical tokens $a_i$, routine syntax and formatting stay near $v_{i,j}\approx 0$, whereas specific factual spans backed by retrieved evidence drop sharply in likelihood once $h_i$ is masked ($v_{i,j}>0$; Figure~\ref{fig:tool-dependency}).
Unlike Self-Aligned Reward (SAR)~\citep{han2025self}, which drops the query and adds a $[-1,1]$ bonus for concise reasoning, \method{} retains $x$ in both branches, deletes only $h_i$, and uses $D_i\in[0,1]$ to gate a search penalty.

\subsection{Correctness-Gated Selective Regularization}
\label{sec:terminal-penalty}
Penalizing failed rollouts discourages search on hard questions, whereas rewarding high $D_i$ invites extra calls and verbatim copying.
Letting $m_i\in\{0,1\}$ mark trajectories with valid Web or Vision search scores (details in Appendix~\ref{sec:dependence-implementation}), we penalize only correct searches below $\theta\in[0,1)$:
\begin{equation}
z_i=\begin{cases}[\theta-D_i]_+,& r_im_i=1,\\0,&r_im_i=0,\end{cases}
\qquad \widetilde r_i=r_i-\beta z_i,\quad \beta\geq0.
\label{eq:penalty}
\end{equation}
Incorrect answers, direct answers, and supported searches ($D_i\ge\theta$) incur zero penalty; when $\beta\theta<1$, every correct rollout keeps $\widetilde r_i\ge 1-\beta\theta>0$ over zero-reward failures.
We set $\beta=1$ and $\theta=0.15$ in our main experiments, requiring at least a $15\%$ drop in geometric-mean token likelihood upon trace removal (Appendix~\ref{sec:run-manifest}).

\subsection{Group-Relative Policy Update}
\label{sec:policy-update}
To credit earlier search decisions without a value critic, we compute the group-normalized GRPO advantage~\citep{shao2024deepseekmath}:
\begin{equation}
\widehat A_i=\frac{\widetilde r_i-\overline{\widetilde r}}
{\operatorname{std}_{j\in\mathcal G}(\widetilde r_j)+\epsilon},
\label{eq:group-advantage}
\end{equation}
where $\epsilon>0$ stabilizes normalization, and the PPO clipped objective~\citep{schulman2017ppo} applies $\widehat A_i$ to generated tokens (excluding observations and padding) while reusing actor weights in no-grad passes without extra memory (Algorithm~\ref{alg:dependence}).
Because $-\beta z_i$ enters through group normalization, Section~\ref{sec:theory} next examines why penalizing low support behaves differently from scaling a cost penalty and how $D_i$ relates to actual tool reliance.
\begingroup
\setlength{\intextsep}{7pt}
\begin{algorithm}[H]
\caption{\method{}: sampling, paired scoring, and policy update.}
\label{alg:dependence}
\small
\begin{algorithmic}[1]
\Require Input $x=(I,q)$, actor $\pi_\phi$, group size $m$, threshold $\theta$, weight $\beta$
\State Sample $\mathcal G=\{\tau_i=(h_i,a_i)\}_{i=1}^{m}\sim\pi_\phi(\cdot\mid x)$; judge $r_i\in\{0,1\}$
\State Set $\widetilde r_i\gets r_i$ ($\forall i$); detach snapshot $\pi_{\bar\phi}\gets\operatorname{stop\_grad}(\pi_\phi)$
\For{each trajectory $i$ with $r_i=1$ and valid tool use ($m_i=1$)}
\State Score fixed $a_i$ under $x\oplus h_i$ and $x$ via Eq.~\eqref{eq:paired-logp} for $\ell_i^{\mathrm{real}},\ell_i^{\mathrm{cf}}$
\State $D_i\gets\operatorname{clip}(1-e^{-(\ell_i^{\mathrm{real}}-\ell_i^{\mathrm{cf}})},0,1)$; $\widetilde r_i\gets r_i-\beta[\theta-D_i]_+$
\EndFor
\State Compute $\{\widehat A_i\}_{i=1}^m$ via Eq.~\eqref{eq:group-advantage}; update $\pi_\phi$ on generated tokens
\end{algorithmic}
\end{algorithm}
\endgroup

\clearpage
\begin{figure}[H]
\centering
\includegraphics[width=0.94\textwidth]{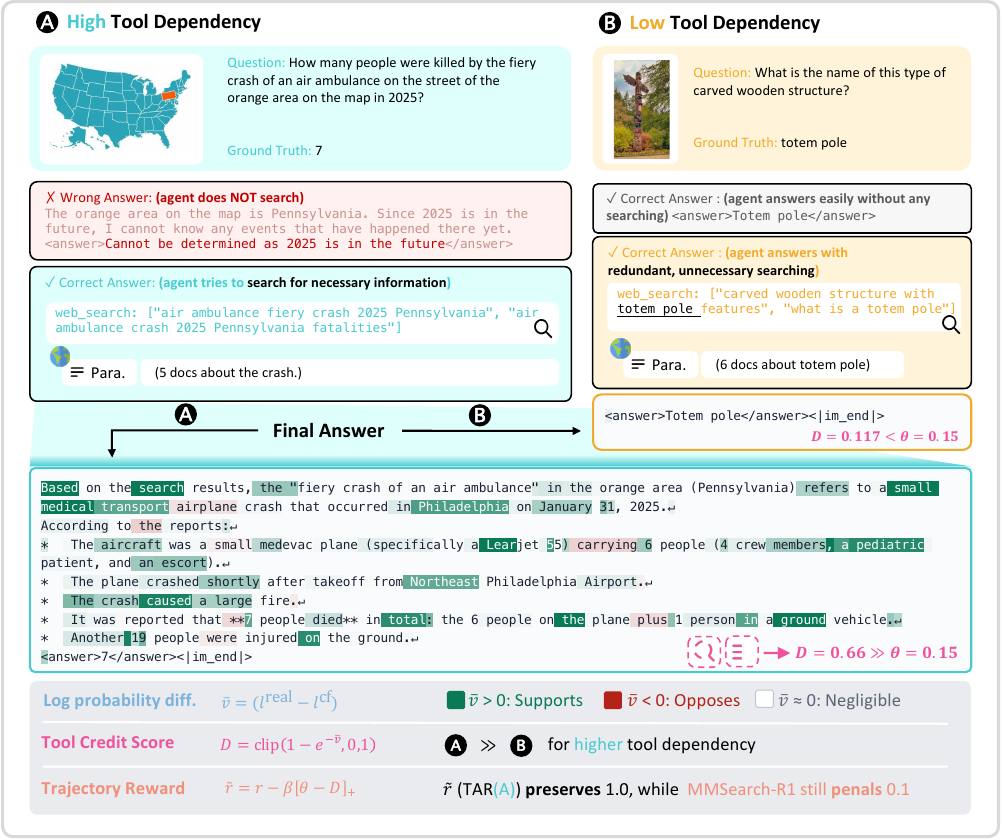}
\caption{%
\textbf{Qualitative comparison of high and low tool dependency under \method{}.}
Token shading indicates positive (\textcolor{capTokenGreen}{green}), negative (\textcolor{capTokenRed}{red}), and negligible (\textcolor{capMediumGray}{gray}) log-probability differences between the full-trace and input-only contexts.
\textbf{(A) High Tool Dependency:} Direct answering fails on a time-sensitive 2025 crash question, whereas web search supplies key factual evidence, yielding a high Tool Credit Score ($D = 0.66 \gg \theta = 0.15$) and preserving the full trajectory reward ($\widetilde r = 1.0$, whereas MMSearch-R1 deducts $0.1$).
\textbf{(B) Low Tool Dependency:} Identifying the totem pole is directly answerable from the image; redundant search yields weak token-level support ($D = 0.117 < \theta = 0.15$) and incurs a \method{} penalty ($\widetilde r = r - \beta[\theta - D]_+$).%
}
\label{fig:tool-dependency}
\end{figure}

\section{Analysis of \method{}}
\label{sec:theory}
We next ask two questions: why support-based penalties act differently from scaling a search cost under GRPO, and how $D_i$ reflects token-level evidence and tool reliance.

\subsection{Penalty Allocation vs. Scaling under GRPO}
\label{sec:count-comparison}
When a cost penalty hurts accuracy, simply shrinking $\beta$ is often ineffective because rescaling $\beta$ and changing which rollouts are penalized ($z_i$) act differently under GRPO normalization (proofs in Appendix~\ref{sec:theory-proofs}).

\begin{lemma}[Penalty-scale cancellation]
\label{lem:penalty-normalization}
If every rollout in a group has reward $r_i=c$ with fixed nonnegative features $z_i$ and $\beta>0$, writing $\sigma_z=\operatorname{std}(z)$ reduces Eq.~\eqref{eq:group-advantage} to
\begin{equation}
\widehat A_i(\beta)=-\frac{z_i-\bar z}{\sigma_z+\epsilon/\beta},
\label{eq:scale-cancellation}
\end{equation}
approaching the $\beta$-independent limit $-(z_i-\bar z)/\sigma_z$ for $\sigma_z>0$ and $\beta\sigma_z\gg\epsilon$.
\end{lemma}
Thus, in all-correct groups ($r_i=1$) with $\beta\sigma_z\gg\epsilon$, lowering a cost weight does not proportionally soften the negative advantage on searched rollouts, explaining why even small baseline penalties hurt accuracy in Figure~\ref{fig:penalty-sweeps}.

\begin{proposition}[Distinguishing equally charged searches]
\label{prop:support-margin}
For two valid correct search trajectories $i,j$ in the same group with equal search cost and $D_i\geq D_j$, letting $s_\beta=\operatorname{std}(\widetilde r)$ gives
\begin{equation}
\widehat A_i-\widehat A_j=
\frac{\beta\bigl([\theta-D_j]_+-[\theta-D_i]_+\bigr)}{s_\beta+\epsilon}\geq0,
\label{eq:support-margin}
\end{equation}
whereas any penalty gated on correctness and search cost alone yields $\widehat A_i-\widehat A_j=0$.
\end{proposition}
More generally, at the pre-clipping on-policy point (excluding KL), with score vector $\mathbf g_i$, $s_0=\operatorname{std}(r)$, and $\mathcal G_0=\sum_{i\in\mathcal G}\widehat A_i(0)\mathbf g_i$, a mixed-group gradient decomposes into
\begin{equation}
\mathcal G_\beta
=\frac{s_0+\epsilon}{s_\beta+\epsilon}\mathcal G_0
-\frac{\beta}{s_\beta+\epsilon}\sum_{i\in\mathcal G}(z_i-\bar z)\mathbf g_i.
\label{eq:gradient-decomposition-main}
\end{equation}
Equation~\eqref{eq:gradient-decomposition-main} shows that \method{} redirects the penalty gradient, concentrating negative weight on low-support searches ($z_i>\bar z$) while sparing well-supported ones ($z_i=0$).
Under expected reward (Appendix~\ref{sec:theory-proofs}), it also lowers the accuracy hurdle for preferring search from $p_1-p_0>\lambda p_1$ to $p_1-p_0>\beta p_1\mu_z$, where $\mu_z=\mathbb E[z\mid r=1,\mathrm{tool}]$.

\subsection{Signal Properties and Behavioral Relevance}
\label{sec:round-support}
Figure~\ref{fig:tool-dependency} contrasts high- and low-dependency cases to show how token-level shifts localize evidence support.
In the high-dependency case~(A), the map identifies Pennsylvania without post-cutoff 2025 crash facts, so answering without search fails ($r=0$).
When web search succeeds ($r=1$), positive shifts ($v_{i,j}>0$, green) concentrate on report-backed facts (city, date, aircraft, and fatality count \texttt{7}), raising $D=0.66\gg\theta=0.15$ and preserving $\widetilde r=1.0$.
In the low-dependency case~(B), the image alone suffices for \texttt{Totem pole} ($r=1$); after a six-document search, masking the trace barely alters token likelihoods on \texttt{<answer>Totem pole</answer>} ($D=0.117<\theta=0.15$), triggering \method{}'s penalty ($\widetilde r=r-\beta[\theta-D]_+$).
By contrast, MMSearch-R1~\citep{wu2025mmsearchr1} still deducts $0.1$ in case~(A) ($\widetilde r=0.9$); per Lemma~\ref{lem:penalty-normalization}, Proposition~\ref{prop:support-margin}, and Eq.~\eqref{eq:gradient-decomposition-main}, such fixed discounts treat needed and redundant searches of equal cost alike and assign a negative advantage whenever an unsearched rollout guesses correctly ($\widetilde r=1.0$).

To test whether teacher-forced likelihood shifts track actual answer reliance, we run a frozen-SFT replay check on 171 (Qwen3.5-35B-A3B) and 151 (Qwen3.5-9B) correct search trajectories, sampling $K=16$ tool-free continuations per branch to measure the accuracy drop $G_i=\widehat p_i^{\mathrm{real}}-\widehat p_i^{\mathrm{cf}}$ upon trace removal (Appendix~\ref{sec:diagnostic-protocol}).
On both backbones, $\bar v_i$ correlates positively with $G_i$ (Spearman $\rho=0.205, 0.252$) and separates $0/16$ no-trace failure cases ($\mathrm{AUC}=0.728, 0.655$).
Meanwhile, rollouts with $D_i<0.15$ still show nonzero mean accuracy drops ($0.550, 0.602$), and across multi-round traces $h_i=B_1\oplus\cdots\oplus B_K$ with prefix log-likelihoods $\ell_{i,k}$, the sum $\bar v_i=\sum_{k=1}^K(\ell_{i,k}-\ell_{i,k-1})$ captures overall trace support rather than individual call contributions (Appendix~\ref{sec:ordered-support}), making $D_i$ suitable as a soft training penalty within GRPO.

\begin{table}[t]
\centering\footnotesize\setlength{\tabcolsep}{2.4pt}\renewcommand{\arraystretch}{1.12}
\caption{%
\textbf{Accuracy and search efficiency on SimpleVQA, DynVQA, and MMSearch.}
Acc.\ (\%), SR (\%), and Tur.\ denote GPT-5.5 judged accuracy, search rate, and generation turns (up to two search rounds plus final answer). MMSearch and Average are weighted by sample count across subtasks and benchmarks, respectively.
Row shading separates resource-penalized RL baselines (\textcolor{capLightGray}{light gray}) from high-accuracy RL methods (\textcolor{capMediumGray}{medium gray} and \textcolor{capPlotBlue}{blue}); \textbf{bold} marks the highest Acc.\ overall and the lowest SR and Tur.\ within the high-accuracy group (Vanilla GRPO, IG-Search, and \method{}).%
}
\label{tab:main}
\resizebox{\linewidth}{!}{%
\begin{tabular}{@{}l*{12}{r}@{}}
\toprule
\multicolumn{13}{c}{\textbf{Qwen3.5-35B-A3B}}\\
\midrule
 & \multicolumn{3}{c}{SimpleVQA} & \multicolumn{3}{c}{DynVQA} & \multicolumn{3}{c}{MMSearch} & \multicolumn{3}{c}{\textbf{Average}}\\
\cmidrule(lr){2-4}\cmidrule(lr){5-7}\cmidrule(lr){8-10}\cmidrule(lr){11-13}
Method & Acc.$\uparrow$ & SR$\downarrow$ & Tur.$\downarrow$ & Acc.$\uparrow$ & SR$\downarrow$ & Tur.$\downarrow$ & Acc.$\uparrow$ & SR$\downarrow$ & Tur.$\downarrow$ & Acc.$\uparrow$ & SR$\downarrow$ & Tur.$\downarrow$\\
\midrule
Base model & 29.48 & 66.02 & 2.32 & 15.60 & 77.50 & 2.54 & 26.92 & 53.51 & 2.05 & 25.47 & 65.16 & 2.29 \\
SFT model & 63.95 & 66.12 & 1.99 & 43.90 & 86.60 & 2.14 & 41.04 & 52.27 & 1.72 & 52.67 & 67.00 & 1.95 \\
\rowcolor{gray!10}
Direct Penalty & 60.84 & 22.47 & 1.23 & 42.90 & 20.00 & 1.22 & 38.56 & 28.16 & 1.35 & 50.24 & 23.51 & 1.26 \\
\rowcolor{gray!10}
SlimSearcher & 54.17 & 22.32 & 1.22 & 28.90 & 16.10 & 1.16 & 32.54 & 16.60 & 1.17 & 42.02 & 19.22 & 1.19 \\
\rowcolor{gray!10}
MMSearch-R1 & 65.43 & 40.40 & 1.59 & 44.00 & 52.00 & 1.79 & 42.03 & 47.98 & 1.82 & 53.68 & 45.30 & 1.70 \\
\rowcolor{gray!20}
Vanilla GRPO & 66.81 & 92.05 & 2.51 & \textbf{45.60} & 99.70 & 2.73 & 43.77 & 97.94 & 2.57 & 55.22 & 95.54 & 2.58 \\
\rowcolor{gray!20}
IG-Search & 67.16 & 91.85 & 2.50 & 45.10 & 99.70 & 2.62 & 43.44 & 97.94 & 2.54 & 55.17 & 95.44 & 2.54 \\
\rowcolor{paperBlueLight}
\textcolor{paperBlue}{\method{}} & \textbf{67.51} & \textbf{68.99} & \textbf{2.35} & \textbf{45.60} & \textbf{91.60} & \textbf{2.43} & \textbf{44.01} & \textbf{74.57} & \textbf{2.28} & \textbf{55.62} & \textbf{75.92} & \textbf{2.35} \\
\midrule[.7pt]
\multicolumn{13}{c}{\textbf{Qwen3.5-9B}}\\
\midrule
 & \multicolumn{3}{c}{SimpleVQA} & \multicolumn{3}{c}{DynVQA} & \multicolumn{3}{c}{MMSearch} & \multicolumn{3}{c}{\textbf{Average}}\\
\cmidrule(lr){2-4}\cmidrule(lr){5-7}\cmidrule(lr){8-10}\cmidrule(lr){11-13}
Method & Acc.$\uparrow$ & SR$\downarrow$ & Tur.$\downarrow$ & Acc.$\uparrow$ & SR$\downarrow$ & Tur.$\downarrow$ & Acc.$\uparrow$ & SR$\downarrow$ & Tur.$\downarrow$ & Acc.$\uparrow$ & SR$\downarrow$ & Tur.$\downarrow$\\
\midrule
Base model & 46.52 & 47.06 & 1.87 & 18.60 & 86.00 & 2.62 & 16.60 & 78.12 & 2.54 & 31.37 & 65.13 & 2.24 \\
SFT model & 60.54 & 64.44 & 1.92 & 38.30 & 79.00 & 2.05 & 37.41 & 51.36 & 1.70 & 48.68 & 64.14 & 1.89 \\
\rowcolor{gray!10}
Direct Penalty & 59.51 & 33.28 & 1.38 & 36.70 & 53.40 & 1.69 & 40.55 & 40.63 & 1.56 & 48.70 & 40.13 & 1.51 \\
\rowcolor{gray!10}
SlimSearcher & 53.19 & 31.06 & 1.36 & 28.60 & 37.50 & 1.48 & 36.75 & 30.88 & 1.45 & 42.68 & 32.53 & 1.42 \\
\rowcolor{gray!10}
MMSearch-R1 & 62.22 & 40.99 & 1.70 & 38.10 & 52.50 & 1.80 & 38.98 & 45.42 & 1.79 & 49.88 & 44.97 & 1.75 \\
\rowcolor{gray!20}
Vanilla GRPO & \textbf{65.48} & 97.78 & 2.64 & 43.90 & 99.90 & 2.68 & 40.63 & 99.26 & 2.61 & \textbf{53.28} & 98.70 & 2.64 \\
\rowcolor{gray!20}
IG-Search & 64.99 & 98.42 & 2.71 & 42.90 & 99.80 & 2.66 & 40.96 & 99.75 & 2.56 & 52.90 & 99.13 & 2.66 \\
\rowcolor{paperBlueLight}
\textcolor{paperBlue}{\method{}} & 64.59 & \textbf{72.49} & \textbf{2.42} & \textbf{44.10} & \textbf{94.10} & \textbf{2.46} & \textbf{41.78} & \textbf{77.29} & \textbf{2.36} & 53.23 & \textbf{78.97} & \textbf{2.41} \\
\bottomrule
\end{tabular}}
\end{table}

\ReportNeedspace{95pt}
\section{Experiments}
\label{sec:exp}
\subsection{Experimental setup}
\label{sec:setup}
\paragraph{Multimodal Search tools.}
To retrieve external context and up-to-date facts, all policies have access to text-based web search (\texttt{web\_search}) and reverse-image search (\texttt{vision\_search}) through Tencent Yuanbao search services.\footnote{\url{https://cloud.tencent.com/product/wsa}}
The agent decides whether to search, which tool to invoke, and what query to issue.
Specifically, \texttt{web\_search} returns relevant documents, and \texttt{vision\_search} queries using the original image and a visual description, without a separate grounding or cropping tool.
Both tools return textual observations rather than additional images while keeping the original input images in context; runtime limits and tool schemas are given in Appendices~\ref{sec:data-manifest} and~\ref{sec:exact-prompts}.

\paragraph{Implementation Details.}
We conduct experiments with Qwen3.5-35B-A3B-Base~\citep{qwen2026base35b} and Qwen3.5-9B-Base~\citep{qwen2026base9b}, abbreviated below as Qwen3.5-35B-A3B and Qwen3.5-9B.
We use ms-swift for supervised fine-tuning and VeRL~\citep{sheng2024hybridflow} for online GRPO training.
Following the SFT-then-online-RL pipeline of DeepMMSearch-R1~\citep{narayan2025deepmmsearch}, we first initialize the policies on 15,802 supervised trajectories from InfoSeek~\citep{chen2023infoseek} and LiveVQA~\citep{fu2025livevqa}, and then optimize them on 4,856 FVQA training questions~\citep{wu2025mmsearchr1}.
All compared RL methods share these training settings and are evaluated at step 50.
Supervised fine-tuning runs for three epochs with a learning rate of $10^{-5}$ and a global batch size of 64, freezing the vision encoder while training the aligner and language model, and uses the epoch-2 checkpoint to initialize RL.
For RL, we use a learning rate of $2\times10^{-6}$ with cosine decay, a KL coefficient of $0.001$, and a clip ratio of $0.2$, sampling 512 prompts with eight rollouts per prompt and a PPO mini-batch of 128 at each step (further details in Appendix~\ref{sec:run-manifest}).

\paragraph{Baselines and Benchmarks.}
\label{sec:baselines}
We compare \method{} against seven baselines across three benchmarks.
Alongside the Qwen3.5 Base, SFT, and Vanilla GRPO models, we evaluate three RL methods designed for search agents: SlimSearcher~\citep{slim2026}, MMSearch-R1~\citep{wu2025mmsearchr1}, and IG-Search~\citep{igsearch2026}, all trained and evaluated within our shared multimodal pipeline.
We also include Direct Penalty, a simple baseline that subtracts a fixed cost per tool call; baseline adaptation and hyperparameter details are provided in Section~\ref{sec:baseline-recipes} of Appendix~\ref{sec:run-manifest}.
Evaluation covers SimpleVQA (2,025)~\citep{cheng2025simplevqa}, DynVQA (1,000)~\citep{liu2024dynvqa}, and MMSearch (1,211)~\citep{jiang2024mmsearch}, totaling 4,236 examples.
Together, these benchmarks span image-grounded factual QA, time-sensitive knowledge retrieval, webpage selection, and multi-source evidence synthesis.
Our MMSearch column pools three 300-example subtask evaluations from MMSearch with the 311-example \emph{MMSearch-Plus} benchmark~\citep{tao2025mmsearchplus}, which requires tracing visual provenance and cross-checking retrieved sources to recover context beyond the image (dataset details in Appendix~\ref{sec:data-manifest}).

\begin{table}[t]
\centering
\caption{%
\textbf{Paired accuracy comparison on no-search decisions.}
Each method is compared with Vanilla GRPO on its own search-omission subset ($N$ records where GRPO searches but the method does not; \% all denotes $N / 4{,}236$).
\textcolor{capMediumGray}{Gray} and \textcolor{capPlotBlue}{blue} bars show Vanilla GRPO and method accuracy (axes start at $20\%$), with hatched \textcolor{capDarkGray}{dark-gray} segments marking GRPO's accuracy advantage where positive.
Direct, Slim, and MMR1 denote Direct Penalty, SlimSearcher, and MMSearch-R1; IG-Search ($N \le 27$) is tabulated but omitted from the plots.%
}
\label{tab:search-omission-counts}
\fontsize{8.5}{10.2}\selectfont
\setlength{\tabcolsep}{0pt}
\renewcommand{\arraystretch}{1.8}
\setlength{\fboxsep}{1pt}
\setlength{\fboxrule}{0.4pt}
\newcommand{\nosearchhead}[1]{\raisebox{4.28pt}[0pt][0pt]{\textbf{#1}}}
\newcommand{\nosearchplot}[1]{%
\fbox{\begin{minipage}[c][88pt][c]{\dimexpr\linewidth-2\fboxsep-2\fboxrule\relax}
\makebox[\linewidth][c]{\hspace*{-3pt}%
\includegraphics[width=\linewidth,height=88pt,keepaspectratio,trim=0 4.476522bp 0 0,clip]{#1}%
\hspace*{3pt}}%
\end{minipage}}}
\begin{tabular}{@{}m{0.118\textwidth}>{\raggedleft\arraybackslash}m{0.043\textwidth}>{\raggedleft\arraybackslash}m{0.065\textwidth}@{\hspace{0.018\textwidth}}>{\centering\arraybackslash}m{0.24\textwidth}@{\hspace{0.032\textwidth}}m{0.118\textwidth}>{\raggedleft\arraybackslash}m{0.043\textwidth}>{\raggedleft\arraybackslash}m{0.065\textwidth}@{\hspace{0.018\textwidth}}>{\centering\arraybackslash}m{0.24\textwidth}@{}}
\toprule
\multicolumn{4}{c}{\nosearchhead{Qwen3.5-35B-A3B}} & \multicolumn{4}{c}{\nosearchhead{Qwen3.5-9B}} \\[-4.08pt]
\cmidrule(lr){1-4}\cmidrule(l){5-8}
\nosearchhead{Method} & \nosearchhead{$N$} & \nosearchhead{\% all} & \nosearchhead{Accuracy comparison} & \nosearchhead{Method} & \nosearchhead{$N$} & \nosearchhead{\% all} & \nosearchhead{Accuracy comparison} \\[-4.08pt]
\midrule
Direct Penalty & 3,051 & 72.03\% & \multirow{5}{=}[0.5pt]{\centering\nosearchplot{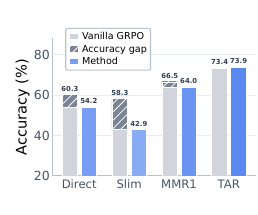}} & Direct Penalty & 2,481 & 58.57\% & \multirow{5}{=}[0.5pt]{\centering\nosearchplot{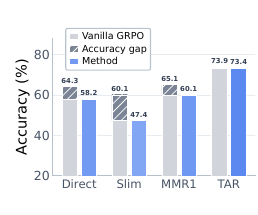}} \\
SlimSearcher & 3,233 & 76.32\% &  & SlimSearcher & 2,803 & 66.17\% &  \\
MMSearch-R1 & 2,128 & 50.24\% &  & MMSearch-R1 & 2,276 & 53.73\% &  \\
IG-Search & 27 & 0.64\% &  & IG-Search & 1 & 0.02\% &  \\
\method{} & 831 & 19.62\% &  & \method{} & 836 & 19.74\% &  \\
\bottomrule
\end{tabular}
\end{table}

\paragraph{Evaluation Metric and Inference.}
Following standard practice in multimodal question answering~\citep{wu2025mmsearchr1,narayan2025deepmmsearch}, we use an LLM-as-Judge setup to evaluate answer accuracy, accommodating semantically equivalent phrasings in open-ended responses that exact string matching misses.
Specifically, GPT-5.5 judges whether each response is correct given the question and reference answers, with full evaluation protocols and prompts in Appendices~\ref{sec:data-manifest} and~\ref{sec:exact-prompts}.
Inference uses greedy decoding with up to 1,024 new tokens per turn across at most three turns, executing the first proposed tool call in each of at most two search rounds.

\subsection{Main Results}
\label{sec:main-results}
Table~\ref{tab:main} reports the main results across all three benchmarks and both backbones; we analyze the key findings below.
\paragraph{Supervised initialization and \method{} substantially improve over the base models.}
The pretrained Qwen3.5 Base models already invoke search on roughly $65\%$ of queries, yet their answer accuracy lags far behind the SFT models.
Supervised initialization raises average accuracy by $27.20\%$ on Qwen3.5-35B-A3B and $17.30\%$ on Qwen3.5-9B while leaving overall search rates nearly unchanged, showing that SFT improves answer and evidence-use quality without simply calling tools more often.
Building on SFT, \method{} brings further accuracy gains alongside a moderate increase in search rate to $75.92\%$ and $78.97\%$, respectively.
Relative to the Base models, \method{} improves average accuracy by $30.15\%$ and $21.86\%$, achieving the highest overall accuracy on Qwen3.5-35B-A3B and coming within $0.05\%$ of the best on Qwen3.5-9B.
Unlike Vanilla GRPO and IG-Search, both of which search on over $95\%$ of queries across both backbones, \method{} pairs strong answer accuracy with selective tool invocation.

\paragraph{\method{} cuts search rate while preserving accuracy.}
Compared with Vanilla GRPO, \method{} reduces search rate on both backbones: on Qwen3.5-35B-A3B, it lowers search rate by $19.62\%$ while raising average accuracy by $0.40\%$ (improving on SimpleVQA and MMSearch and matching on DynVQA), and on Qwen3.5-9B, it lowers search rate by $19.74\%$ with only a $0.05\%$ difference in average accuracy.
Notably, on MMSearch, \method{} achieves the highest overall accuracy on both backbones, cutting search rate by $21.97\%$ (from $99.26\%$ to $77.29\%$) while improving accuracy by $1.15\%$ on Qwen3.5-9B.
This reduction is concentrated in ReRank and Summarization, where candidate websites or webpage text are already supplied in the prompt, whereas End-to-End and MMSearch-Plus retain high search rates for open-ended evidence gathering (subtask results in Appendix~\ref{sec:main5-audit}).
Mean generation turns likewise drop on both backbones, whereas IG-Search continues to search on nearly every query.
This contrast stems from what each method rewards: IG-Search credits queries that raise gold-answer likelihood over random documents~\citep{igsearch2026}, which improves query quality without discouraging unnecessary search initiation.
\method{} instead compares response likelihood with and without the interaction trace and penalizes only correct searches whose support falls below $\theta$, leaving direct answers and informative searches unpenalized so redundant calls drop without hurting accuracy.

\paragraph{Penalizing correct searches without measuring support hurts accuracy.}
Direct Penalty, SlimSearcher, and MMSearch-R1 all search less often than the SFT model, but do so at a clear cost in answer accuracy.
This happens because Direct Penalty and MMSearch-R1 penalize every correct tool-using trajectory (including queries that genuinely require retrieval, as in Figure~\ref{fig:tool-dependency}(A)), while SlimSearcher discounts higher-cost or longer rollouts relative to the group's lowest-cost correct rollout without checking whether the trace supports the answer.
As Section~\ref{sec:analysis} shows, even small penalty weights reduce both search rate and overall accuracy for these baselines.

\subsection{Extended Analysis}
\label{sec:analysis}\label{sec:behavior}
\paragraph{Accuracy and coverage of no-search decisions.}
Table~\ref{tab:search-omission-counts} compares each method's answer accuracy with Vanilla GRPO on the subset of questions where Vanilla GRPO searches but the method does not.
\method{} skips search on roughly one fifth of the queries where Vanilla GRPO searches, compared with more than half for each cost-penalized baseline.
\method{} shows the smallest paired accuracy gap among all plotted methods, scoring $0.48\%$ above Vanilla GRPO on Qwen3.5-35B-A3B and $0.48\%$ below on Qwen3.5-9B.
It answers over $73\%$ of each subset correctly without searching, matching the removal of low-support searches illustrated in Figure~\ref{fig:tool-dependency}(B), whereas other baselines suffer much steeper drops (four-state breakdown and shared-slice controls in Appendix~\ref{sec:search-state-decomposition}).

\begin{figure}[!htbp]
\centering
\begin{subfigure}[t]{0.45\columnwidth}
  \centering
  \includegraphics[width=\linewidth]{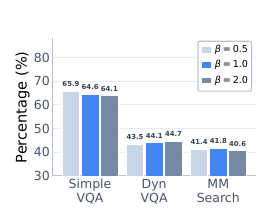}
  \caption{Accuracy}
  \label{fig:beta-sensitivity-acc}
\end{subfigure}\hspace{.035\linewidth}
\begin{subfigure}[t]{0.45\columnwidth}
  \centering
  \includegraphics[width=\linewidth]{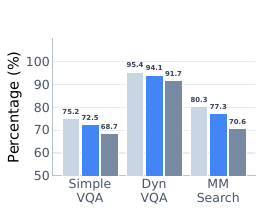}
  \caption{Search rate}
  \label{fig:beta-sensitivity-sr}
\end{subfigure}
\caption{%
\textbf{\method{} coefficient sensitivity on Qwen3.5-9B.}
Panels compare per-benchmark accuracy and search rate across penalty weights $\beta \in \{0.5, 1.0, 2.0\}$ ($\beta = 1.0$ in Table~\ref{tab:main}); larger $\beta$ monotonically reduces search rate on all three benchmarks with minor accuracy variation.%
}
\label{fig:beta-sensitivity}
\end{figure}

\paragraph{\method{} coefficient sensitivity and search control.}
\label{sec:sensitivity}
Figure~\ref{fig:beta-sensitivity} examines how $\beta \in \{0.5, 1.0, 2.0\}$ controls the accuracy--search trade-off on Qwen3.5-9B across all 4,236 examples.
At $\beta=0.5$, \method{} raises average accuracy over Vanilla GRPO by $0.31\%$ (from gains on SimpleVQA and MMSearch, with a small drop on DynVQA) while cutting search rate by $17.28\%$.
Increasing $\beta$ from $1.0$ to $2.0$ cuts search rate by a further $4.29\%$ with only a $0.42\%$ drop in average accuracy, providing smooth control over search frequency around the default $\beta = 1.0$ reported in Table~\ref{tab:main} (protocol details in Appendix~\ref{sec:tar-sensitivity-controls}).

\begin{figure}[!htbp]
\centering
\begin{subfigure}[t]{0.43\columnwidth}
  \centering
  \includegraphics[width=\linewidth]{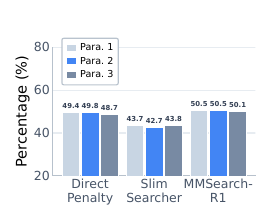}
  \caption{Overall accuracy}
  \label{fig:baseline-sensitivity-acc}
\end{subfigure}\hspace{.035\linewidth}
\begin{subfigure}[t]{0.43\columnwidth}
  \centering
  \includegraphics[width=\linewidth]{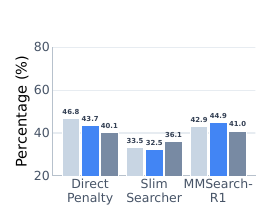}
  \caption{Search rate}
  \label{fig:baseline-sensitivity-sr}
\end{subfigure}
\caption{%
\textbf{Baseline parameter sensitivity on Qwen3.5-9B.}
Para.~1--3 denote $c = 0.01, 0.10, {0.15}$ for Direct Penalty, $\lambda_C = 0.3, {1.0}, 3.0$ for SlimSearcher, and $p = 0.05, {0.10}, 0.30$ for MMSearch-R1 (bold indicates Table~\ref{tab:main} defaults).
All settings pool the full 4,236-record population (including all four MMSearch subtasks) by sample count, where Vanilla GRPO attains $53.28\%$ accuracy and a $98.70\%$ search rate; both vertical axes start at $20\%$.%
}
\label{fig:penalty-sweeps}
\end{figure}

\paragraph{Baseline parameter sensitivity.}
\label{sec:baseline-sensitivity}
To examine the accuracy loss discussed in Section~\ref{sec:main-results}, Figure~\ref{fig:penalty-sweeps} compares three penalty settings per baseline on Qwen3.5-9B at step 50 across all 4,236 examples.
Even at the smallest tested penalties (Para.~1), search rates drop from $98.70\%$ under Vanilla GRPO to $33.45\%$--$46.81\%$, while accuracy falls from $53.28\%$ to $43.67\%$--$50.54\%$.
The accuracy drop thus appears even under weak penalties rather than only under strong regularization.
Across all tested settings, the best accuracies of Direct Penalty, SlimSearcher, and MMSearch-R1 remain $3.47\%$, $9.51\%$, and $2.74\%$ below Vanilla GRPO, respectively.
These results confirm Section~\ref{sec:theory} and Figure~\ref{fig:tool-dependency}(A): shrinking a cost-only penalty weight still penalizes helpful searches, showing why \method{}'s support-based penalty assignment is needed (per-task results in Appendix~\ref{sec:direct-penalty-sweeps}).

\FloatBarrier
\section{Conclusion}
\label{sec:conclusion}
We presented Tool-Attributed Regularization (\method{}) to train multimodal agents to use external search selectively.
Instead of charging tool calls by cost alone, \method{} uses the agent itself to compare the likelihood of a completed response with and without its intermediate interaction trace under the same multimodal input.
From this likelihood difference, the Tool Credit Score measures how strongly the trace supports the final answer.
During GRPO training, \method{} penalizes only correct searches with weak support and adds no extra model or scoring cost at inference time.
Our analysis shows why rescaling a cost-only penalty cannot separate helpful from redundant searches of equal cost under GRPO normalization, whereas support-based penalties distinguish their advantages within each rollout group.
Across two Qwen3.5 backbones and three multimodal search benchmarks, \method{} cuts search rate by nearly $20\%$ relative to Vanilla GRPO while keeping average accuracy within $0.5\%$.
It also preserves accuracy when skipping search, controls search rate smoothly across penalty weights, and outperforms existing cost-penalized baselines in accuracy.

Two extensions follow naturally from our current setup.
First, because \method{} removes the intermediate trace as a whole to measure overall support, refining this comparison into turn-level attribution could provide finer credit assignment across longer multi-step tool chains.
Second, while our evaluation focuses on web and reverse-image search returning text, adapting trace-support regularization to visual tools that return new image views, such as region cropping, is a promising next step.

\clearpage
\pdfbookmark[1]{References}{report.references}
\bibliographystyle{styles/iclr2027/iclr2027_conference}
\bibliography{main}
\clearpage
\appendix
\setcounter{figure}{0}
\setcounter{table}{0}
\setcounter{equation}{0}
\renewcommand{\thefigure}{S\arabic{figure}}
\renewcommand{\thetable}{S\arabic{table}}
\renewcommand{\theequation}{S\arabic{equation}}
\renewcommand{\theHfigure}{supp.\arabic{figure}}
\renewcommand{\theHtable}{supp.\arabic{table}}
\renewcommand{\theHequation}{supp.\arabic{equation}}

\addtocontents{toc}{\protect\setcounter{tocdepth}{2}}
\clearpage
\phantomsection
\pdfbookmark[0]{Appendix}{appendix.contents}
\begin{ReportContents}
\begingroup
\centering
{\ReportHeadingFont\fontsize{12}{14}\selectfont Appendix\par}
\endgroup
\vspace{4pt}
\begingroup
\hypersetup{linkcolor=reportTencentBlue,linktoc=all}
\makeatletter
\renewcommand*{\l@section}[2]{%
  \addvspace{7pt}%
  {\bfseries\@dottedtocline{1}{0em}{1.8em}{#1}{\normalfont #2}}}
\renewcommand*{\l@subsection}{\@dottedtocline{2}{1.8em}{2.4em}}
\renewcommand*{\@dotsep}{2.5}
\setlength{\parskip}{3pt}
\edef\appendixsavedtocdepth{\number\value{tocdepth}}
\@starttoc{toc}
\setcounter{tocdepth}{\appendixsavedtocdepth}
\makeatother
\endgroup
\end{ReportContents}
\clearpage

\definecolor{tableModelGray}{gray}{0.92}
\definecolor{tableOursBlue}{HTML}{CFEFFF}

\renewcommand{\arraystretch}{1.15}
\setlength{\tabcolsep}{5pt}
\setlength{\LTpre}{6pt}
\setlength{\LTpost}{8pt}
\newcommand{\suppgroup}[2]{\rowcolor{tableModelGray}\multicolumn{#1}{c}{\rule{0pt}{2.4ex}\textbf{#2}}\\}

\section{Implementation Details}
\label{sec:dependence-implementation}
This section details the paired forward scoring pass and terminal reward update in Algorithm~\ref{alg:dependence}, following the notation of Section~\ref{sec:method}.

\textbf{Final response extraction and gradient masking.} For each completed rollout, we locate the final contiguous block of assistant-generated tokens $a_i$ by scanning the response mask backward past any trailing padding.
This segment (comprising concluding reasoning, answer tags, and the end-of-sequence token) is scored identically under both contexts, whereas semantic answer extraction is handled separately by the correctness judge.
During the GRPO policy update, tool observations and padding tokens are masked out so gradients flow solely through assistant-generated tokens across all turns.

\textbf{Paired context construction.} We construct the full-trace sequence $C_i^{\mathrm{real}}\oplus a_i=(x\oplus h_i)\oplus a_i$ and the input-only sequence $C_i^{\mathrm{cf}}\oplus a_i=x\oplus a_i$, where $h_i$ collects all intermediate reasoning, tool calls, and returned textual observations.
Both branches retain the original input images and evaluate the exact same token IDs under a detached actor snapshot without resampling, rebuilding contiguous positional indices for the shorter input-only prefix.

\textbf{Eligibility, validity checks, and terminal penalty placement.} Only correct, non-padding trajectories that execute at least one search call ($e_i = r_i m_i = 1$) enter the \method{} penalty computation.
Both forward passes must return nonempty, equal-length, finite log-probability vectors; if pair construction fails, or if all eligible pairs in a batch yield identically zero differences (guarding against context misalignment), those trajectories retain their unmodified outcome reward.
For valid eligible trajectories, the detached scalar penalty $P_i = \beta[\theta - D_i]_+$ is subtracted once at the final generated token $t_i^\star$ prior to group normalization.

\section{Training Data and Optimization Details}
\label{sec:run-manifest}
Table~\ref{tab:configuration} summarizes the shared training and evaluation settings across all compared RL methods, with baseline-specific reward formulations given in Section~\ref{sec:baseline-recipes}.

\textbf{Supervised initialization (SFT).} To establish a common starting policy capable of multi-turn multimodal search, we synthesize 15,802 training trajectories from InfoSeek~\citep{chen2023infoseek} and LiveVQA~\citep{fu2025livevqa} using a post-trained Qwen3.5-397B-A17B teacher.
We filter trajectories for answer correctness via exact-match gating, remove duplicates against LiveVQA text pairs and FVQA-overlapping items, and compute cross-entropy loss solely on assistant-generated tokens (masking system prompts, user inputs, image tokens, and tool observations).
Both backbones train for three epochs, and the epoch-2 checkpoint serves as the shared initialization for all RL methods.

\textbf{Online GRPO training.} Starting from this shared SFT checkpoint, every RL method runs for 50 updates on 4,856 FVQA~\citep{wu2025mmsearchr1} questions in VeRL~\citep{sheng2024hybridflow} GRPO under the shared hyperparameters in Table~\ref{tab:configuration}.
Each rollout permits up to three assistant turns and two search rounds, executing at most one tool call per round; training-time correctness is judged by Qwen3.5-397B-A17B at temperature $0$ using the prompts in Appendix~\ref{sec:exact-prompts}.

\begin{table}[!htbp]
\centering\small
\setlength{\tabcolsep}{6pt}
\caption{%
\textbf{Summary of shared training and evaluation settings.}
All compared RL methods share the same SFT initialization, FVQA training pool, optimization hyperparameters, and evaluation environment.%
}
\label{tab:configuration}
\begin{tabular}{@{}p{.23\textwidth}p{.73\textwidth}@{}}
\textbf{Component} & \textbf{Configuration}\\
\midrule
Backbones & Qwen3.5-35B-A3B (MoE) and Qwen3.5-9B (dense), initialized from Base checkpoints.\\
SFT data & 15,802 InfoSeek~\citep{chen2023infoseek} and LiveVQA~\citep{fu2025livevqa} trajectories synthesized by Qwen3.5-397B-A17B.\\
SFT optimization & ms-swift; frozen vision encoder; bf16 and ZeRO-3; learning rate $10^{-5}$, global batch size 64, max sequence length 10,240; 3 epochs (epoch-2 checkpoint used to initialize RL).\\
RL data and budget & 4,856 FVQA~\citep{wu2025mmsearchr1} training questions; 50 policy updates.\\
RL optimization & VeRL~\citep{sheng2024hybridflow} GRPO; FSDP2; 512 prompts $\times$ 8 rollouts per update; PPO mini-batch 128; 1 PPO epoch; learning rate $2\times10^{-6}$ with cosine decay; KL coefficient $0.001$; clip ratio $0.2$; frozen vision encoder; prompt/response limits 4,096/10,240 tokens; seed 42.\\
\rowcolor{tableOursBlue}\method{} regularization & Web + Vision tool eligibility, $\beta = 1.0$, $\theta = 0.15$; detached same-actor paired forward passes, zero inference overhead.\\
Inference decoding & Greedy decoding (temperature $0$, top-$p$ $1$), up to 1,024 new tokens per turn, at most 3 assistant turns and 2 search rounds (first parsed call executed per round).\\
Search tools & \texttt{web\_search} (up to 5 documents) and \texttt{vision\_search} (up to 2 documents), returning text observations truncated at 32,768 characters while retaining original input images.\\
Evaluation panel & SimpleVQA~\citep{cheng2025simplevqa} (2,025), DynVQA~\citep{liu2024dynvqa} (1,000), and MMSearch~\citep{jiang2024mmsearch,tao2025mmsearchplus} (1,211 across four subtasks); 4,236 examples in total.\\
Evaluation judge & GPT-5.5 LLM-as-Judge (prompts in Appendix~\ref{sec:exact-prompts}), up to 768 output tokens.\\
\end{tabular}
\end{table}

\subsection{Baseline Details}
\label{sec:baseline-recipes}
All compared RL baselines start from the exact same Qwen3.5-35B-A3B and Qwen3.5-9B SFT checkpoints and share the same rollout budget, tool environment, and optimizer settings as \method{}, differing solely in reward shaping.

\textbf{Vanilla GRPO} optimizes binary answer correctness $r_i \in \{0, 1\}$ alone without any tool-call penalty, serving as the unregularized online RL reference.

\textbf{Direct Penalty} applies a correctness-gated per-query cost, $r_i^{\mathrm{Direct}} = r_i(1 - c\,Q_i)$, where $Q_i$ counts search queries issued across the trajectory (a single tool invocation may contain multiple queries; default $c = 0.15$, with a 9B sweep over $c \in \{0.01, 0.10, 0.15\}$).

\textbf{MMSearch-R1}~\citep{wu2025mmsearchr1} applies a correctness-gated occurrence discount whenever any tool is invoked, $r_i^{\mathrm{MMR1}} = r_i(1 - p\,\mathbb{I}[N_i > 0])$ (default $p = 0.10$, with a 9B sweep over $p \in \{0.05, 0.10, 0.30\}$).

\textbf{SlimSearcher}~\citep{slim2026} scales rewards among successful rollouts within each group according to relative tool cost and response length (default $\lambda_C = 1.0$, with a 9B sweep over $\lambda_C \in \{0.3, 1.0, 3.0\}$).

\textbf{IG-Search}~\citep{igsearch2026} assigns query-level information-gain credit by comparing reference-answer likelihood under retrieved documents versus random documents within our multimodal search environment.
In our codebase, Direct Penalty uses a cost-scaled advantage denominator, whereas the remaining methods use the standard group standard-deviation normalization in Eq.~\eqref{eq:group-advantage}.

\section{Proofs and Derivations}
\label{sec:theory-proofs}
Throughout this section, we consider a fixed rollout group with detached support scores and fixed eligibility masks, using population standard deviations with $\epsilon > 0$ added outside the standard deviation as in Eq.~\eqref{eq:group-advantage}.

\subsection{Proof of Lemma~\ref{lem:penalty-normalization} (Penalty-Scale Cancellation)}
\label{sec:penalty-scale-proof}
\begin{proof}
With constant base reward $r_i = c$ and regularized reward $\widetilde r_i = c - \beta z_i$ ($\beta > 0$), centering gives $\widetilde r_i - \overline{\widetilde r} = -\beta(z_i - \bar z)$ and positive homogeneity gives $\operatorname{std}(\widetilde r) = \beta \sigma_z$.
Dividing the numerator and denominator of Eq.~\eqref{eq:group-advantage} by $\beta$ yields Eq.~\eqref{eq:scale-cancellation}:
\begin{equation}
\widehat A_i(\beta) = -\frac{z_i - \bar z}{\sigma_z + \epsilon/\beta}.
\end{equation}
For $\sigma_z > 0$ and $\beta\sigma_z \gg \epsilon$, this expression approaches the $\beta$-independent limit $-(z_i - \bar z)/\sigma_z$; when $z_i$ is constant across the group, all centered advantages vanish, and for fixed $\epsilon > 0$ they also vanish as $\beta \to 0$.
\end{proof}

This cancellation holds specifically within equal-base-reward groups under standard group normalization; in mixed-correctness groups or under cost-scaled denominators, the effective scale varies with the combined group variance.

\subsection{Proof of Proposition~\ref{prop:support-margin} and Gradient Decomposition}
\label{sec:support-margin-proof}
\begin{proof}[Proof of Proposition~\ref{prop:support-margin}]
For two valid correct search trajectories $i, j \in \mathcal G$, the regularized rewards are $\widetilde r_k = 1 - \beta[\theta - D_k]_+$.
Subtracting their normalized advantages cancels the group mean $\overline{\widetilde r}$ and gives Eq.~\eqref{eq:support-margin}:
\begin{equation}
\widehat A_i - \widehat A_j = \frac{\beta\bigl([\theta - D_j]_+ - [\theta - D_i]_+\bigr)}{s_\beta + \epsilon} \ge 0
\end{equation}
whenever $D_i \ge D_j$, with strict inequality when $\beta > 0$ and the two hinge values differ; conversely, any penalty determined solely by correctness and search cost assigns identical rewards to equally costly correct rollouts and yields $\widehat A_i - \widehat A_j = 0$.
\end{proof}

\textbf{Relative ordering and advantage sign.} A correct search rollout receives a negative advantage precisely when its regularized reward falls below the group mean $\overline{\widetilde r}$.
In an all-correct group containing at least one unpenalized direct answer, uniformly discounted searches always fall below the mean, whereas in mixed-correctness groups a penalized correct search can still retain a positive advantage while obeying the pairwise ordering in Proposition~\ref{prop:support-margin}.

\textbf{Mixed-group gradient decomposition.} At the pre-clipping on-policy point (excluding KL), let $\mathbf g_i$ denote the generated-token policy-score vector for trajectory $i$, and write $\mathcal G_\beta = \sum_{i\in\mathcal G}\widehat A_i(\beta)\mathbf g_i$ with $s_0 = \operatorname{std}(r)$ and $s_\beta = \operatorname{std}(r - \beta z)$.
Expanding the centered reward as $(r_i - \bar r) - \beta(z_i - \bar z)$ yields Eq.~\eqref{eq:gradient-decomposition-main}:
\begin{equation*}
\mathcal G_\beta
= \frac{s_0 + \epsilon}{s_\beta + \epsilon}\,\mathcal G_0
- \frac{\beta}{s_\beta + \epsilon}\sum_{i\in\mathcal G}(z_i - \bar z)\mathbf g_i.
\end{equation*}
The second term reallocates trajectory weights within the group: weakly supported searches ($z_i > \bar z$) receive negative weight, whereas zero-hinge supported searches ($z_i = 0 \le \bar z$) gain relative weight through group centering.

\subsection{Multi-Round Trace Support Decomposition}
\label{sec:ordered-support}
Dropping trajectory index $i$ for brevity, partition a multi-round trace in interaction order as $h = B_1 \oplus \cdots \oplus B_K$, where block $B_k$ contains the reasoning, tool call, and returned observation of round $k$.
Let $h_{\le 0} = \varnothing$, $h_{\le k} = B_1 \oplus \cdots \oplus B_k$, and $C_k = x \oplus h_{\le k}$; under the same detached scorer $\pi_{\bar\phi}$ and fixed response tokens $a = (a_1,\ldots,a_L)$, define the prefix log-likelihood and round increment:
\begin{equation}
\ell_k = \frac{1}{L}\sum_{j=1}^L \log \pi_{\bar\phi}(a_j \mid C_k, a_{<j}), \qquad
\delta_k = \ell_k - \ell_{k-1},
\label{eq:prefix-scores}
\end{equation}
so that $\ell_0 = \ell^{\mathrm{cf}}$ and $\ell_K = \ell^{\mathrm{real}}$.

Telescoping gives $\bar v = \ell_K - \ell_0 = \sum_{k=1}^{K}\delta_k$.
To check whether opposing round increments cancel in practice, we evaluate these intermediate prefixes via teacher forcing across all 151 frozen Qwen3.5-9B diagnostic trajectories (Appendix~\ref{sec:diagnostic-protocol}).
All 151 trajectories satisfy $\sum_{k=1}^{K}\delta_k = \bar v$ (within a maximum numerical-precision discrepancy of $0.0119$ against archived totals) and show negligible inter-round cancellation ($\kappa = 1 - |\bar v|/\sum_{k=1}^K |\delta_k| \approx 0$), confirming that whole-trace likelihood shifts maintain a consistent sign across rounds.

\subsection{Two-Action Expected-Reward Threshold}
\label{sec:two-action-model}
Consider a query with direct-answering accuracy $p_0$ and search accuracy $p_1$, and let $\mu_z = \mathbb E[z \mid r=1, \mathrm{tool}] \in [0, \theta]$ denote the expected \method{} hinge on successful searches (with $\mu_z = 0$ if $p_1 = 0$).
The expected rewards of direct answering, a uniform correctness-gated discount $\lambda$, and \method{} are
\begin{equation}
U_0 = p_0, \qquad
U_1^{\mathrm{uniform}} = p_1(1 - \lambda), \qquad
U_1^{\mathrm{\method{}}} = p_1(1 - \beta \mu_z).
\end{equation}
Searching yields strictly higher expected reward than direct answering whenever $p_1 - p_0 > \lambda p_1$ under a uniform discount, compared with $p_1 - p_0 > \beta p_1 \mu_z$ under \method{}.
Whenever $\beta\mu_z \le \lambda$ (including the matched maximum-penalty setting $\lambda = \beta\theta$), \method{} imposes a lower accuracy-gain threshold for retaining useful search.

\section{Frozen-Policy Behavioral Diagnostics}
\label{sec:diagnostic-protocol}\label{sec:mechanism}
To test whether the teacher-forced log-likelihood shift $\bar v_i$ tracks actual reliance on retrieved evidence, we run a continuation replay diagnostic on originally correct search trajectories from the frozen SFT policies: 171 trajectories (170 image clusters) for Qwen3.5-35B-A3B and 151 trajectories (149 image clusters) for Qwen3.5-9B.
For each trajectory, we remove the final answering segment and sample $K = 16$ tool-free continuations from both the full-trace context $C_i^{\mathrm{real}}$ and the input-only context $C_i^{\mathrm{cf}}$, judging every continuation with Qwen3.5-397B-A17B.
The difference in continuation accuracy defines the accuracy drop upon trace removal, $G_i = \widehat p_i^{\mathrm{real}} - \widehat p_i^{\mathrm{cf}}$, with 95\% ranges computed via 5,000 image-cluster resamples.

\begin{figure}[!htbp]
\centering
\includegraphics[width=\linewidth]{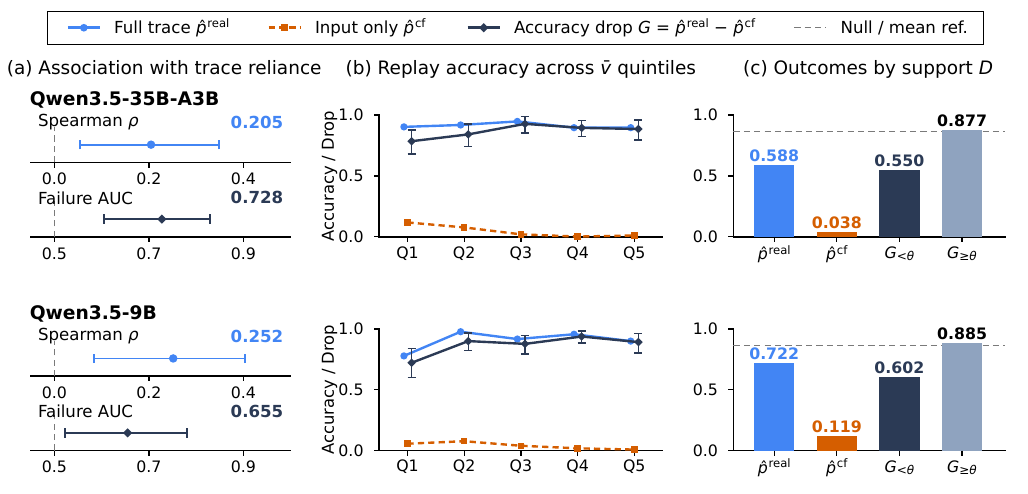}
\caption{%
\textbf{Frozen-policy replay diagnostic across Qwen3.5-35B-A3B and Qwen3.5-9B.}
(a)~Spearman rank correlation $\rho(\bar{v}, G)$ (dashed line at $0.0$) and ROC AUC for identifying $0/16$ input-only replay failures (dashed line at $0.5$).
(b)~Mean replay accuracy under full-trace ($\widehat{p}^{\mathrm{real}}$) and input-only ($\widehat{p}^{\mathrm{cf}}$) contexts alongside accuracy drop $G = \widehat{p}^{\mathrm{real}} - \widehat{p}^{\mathrm{cf}}$ across within-model $\bar{v}$ quintiles.
(c)~Outcomes on low-support trajectories ($D < 0.15$: $\widehat{p}^{\mathrm{real}}$, $\widehat{p}^{\mathrm{cf}}$, and $G_{<\theta}$) compared against the remaining trajectories ($G_{\ge\theta}$), with dashed horizontal lines marking overall mean $G$; error bars show 95\% ranges across 5,000 image-cluster resamples.%
}
\label{fig:dependence}
\end{figure}

\textbf{Likelihood shift $\bar v_i$ tracks continuation accuracy drop upon trace removal.}
As shown in Figure~\ref{fig:dependence}(a--b), $\bar v_i$ correlates positively with accuracy drop $G_i$ on both backbones ($\rho = 0.205\;[0.054, 0.349]$ on 35B-A3B and $\rho = 0.252\;[0.085, 0.404]$ on 9B, with $147/171$ and $131/151$ zero-success input-only rollouts) and separates $0/16$ input-only failures ($\mathrm{AUC} = 0.728\;[0.605, 0.830]$ and $0.655\;[0.522, 0.781]$).
Below the penalty threshold ($D < 0.15$; Figure~\ref{fig:dependence}(c)), trajectories exhibit smaller accuracy drops upon trace removal than the remaining trajectories: $-0.283$ lower on Qwen3.5-9B ($0.602$ vs.\ $0.885$, $n=11$ vs.\ $140$, 95\% range $[-0.516, -0.067]$) and $-0.327$ lower on Qwen3.5-35B-A3B ($0.550$ vs.\ $0.877$; because only $n=5$ of $171$ frozen-SFT trajectories fall below $0.15$, the small-sample range spans $[-0.812, +0.052]$).
At the same time, masking the full intermediate trace still lowers continuation accuracy on $D < 0.15$ rollouts (mean $G = 0.550$ and $0.602$), motivating the use of $D$ as a bounded soft penalty within GRPO rather than a hard call-pruning rule.

\textbf{Comparison with input-only likelihood.} Because correct search rollouts generally assign high log-likelihood $\ell^{\mathrm{real}}$ under the full trace, cross-trajectory variation in $\bar v = \ell^{\mathrm{real}} - \ell^{\mathrm{cf}}$ is largely driven by the input-only log-likelihood $\ell^{\mathrm{cf}}$ (how plausible the completed answer already is from the image and question alone).
Under five-fold image-grouped out-of-fold fitting, an input-only predictor attains zero-success AUCs of $0.509$ (35B-A3B) and $0.634$ (9B), compared with $0.534$ and $0.610$ for $\bar v$ and $0.484$ and $0.636$ for a joint two-feature fit (distinct from the raw-score AUCs in Figure~\ref{fig:dependence}(a)).
Consistent with this observation, rescaling $-\ell^{\mathrm{cf}}$ via empirical dataset statistics achieves a comparable accuracy--search trade-off when used for 9B RL training in Appendix~\ref{sec:tar-sensitivity-controls}, whereas \method{} uses $\ell^{\mathrm{real}}$ as a self-calibrating reference on the same response so $D_i \in [0,1]$ requires no pre-fitted normalization constants.

\textbf{Paired search-blocking check.} To measure the local utility of tool calls on the frozen Qwen3.5-9B policy, we run a paired allow-versus-block rollout intervention from identical pre-call decision prefixes (300 matched pairs per setting).
Blocking the first tool call lowers GPT-5.5 accuracy from $45.67\%$ to $38.67\%$ ($-7.00\%$), while blocking a randomly chosen call lowers accuracy from $44.67\%$ to $39.67\%$ ($-5.00\%$); both interventions flip final correctness on $17\%$ of pairs and trigger an average of $0.27$ compensatory follow-up calls per blocked rollout (earlier checks disabling tools after the first prefix similarly lowered accuracy from $55.62\%$ to $46.24\%$ on 35B-A3B and from $53.23\%$ to $43.53\%$ on 9B).

\section{Evaluation Benchmarks and Judging Protocol}
\label{sec:data-manifest}
\textbf{Benchmarks and split separation.} Our evaluation panel comprises SimpleVQA~\citep{cheng2025simplevqa} (2,025 examples), DynVQA~\citep{liu2024dynvqa} (1,000 examples; February 2025 snapshot), and MMSearch~\citep{jiang2024mmsearch} (1,211 examples: 300 each for End-to-End, ReRank, and Summarization, plus 311 from MMSearch-Plus~\citep{tao2025mmsearchplus}), totaling 4,236 examples.
These evaluation benchmarks are disjoint at the dataset level from both the SFT sources (InfoSeek and LiveVQA) and the RL training pool (FVQA).

\textbf{Agent inference and tool execution.} All policies, including Base and SFT, are evaluated in agent mode with both \texttt{web\_search} and \texttt{vision\_search} enabled.
Decoding is greedy (temperature $0$, top-$p$ $1$) with up to 1,024 new tokens per turn across at most three assistant turns and two search rounds, executing the first parsed tool call per round.
Both tools return textual observations truncated at 32,768 characters while keeping all user input images in context, returning up to 5 documents for \texttt{web\_search} and up to 2 matched entries for \texttt{vision\_search}.

\textbf{LLM-as-Judge protocol and field checks.} Answer accuracy is evaluated by GPT-5.5 using the prompts in Appendix~\ref{sec:exact-prompts}, which receive the image, question, reference answers, and extracted response (defaulting to the full final message when answer tags are absent); content-filtered requests fall back first to text-only judging and then to deterministic matching.
In protocol field audits, passing multiple-choice options in the candidate field interacts with Rule~10's any-candidate matching, and ReRank resolves shuffled candidate indices against stored labels: rejudging fixed responses flipped $253$ labels from incorrect to correct and $402$ from correct to incorrect across $4{,}422$ audited responses, and shifted ReRank accuracy from $41.69\%$ to $38.31\%$ across $4{,}200$ responses (while earlier option-permutation checks showed $0\%$ net change); because all methods in Table~\ref{tab:main} were scored under the identical archived pipeline, we retain the unified archived judgments throughout for consistent cross-method comparison.

\textbf{Sample-weighted aggregation.} All aggregate metrics weight individual examples equally: MMSearch pools its 1,211 examples and Average/Overall pools all 4,236 examples, aggregating accuracy, search rate, tool calls, turns, and tokens from unrounded per-example records before display rounding.

\section{Prompts and Tool Schemas}
\label{sec:exact-prompts}
The boxes below (\hyperref[lst:actor-system]{Prompts~1}--\hyperref[lst:tool-schema]{4}) reproduce the exact agent system prompt, LLM-as-Judge templates, and tool function schemas used across training, diagnostics, and evaluation.
In the actor's user message, input images precede the question text (with multiple-choice options appended in dataset order when present); ReRank supplies eight candidate webpages (index, title, snippet, and URL) directly in the prompt, and Summarization provides up to 16,000 characters of webpage text to the actor (and up to 8,000 characters to the judge).

\par\vspace{2pt}
\phantomsection\label{lst:actor-system}
\includegraphics[width=\linewidth]{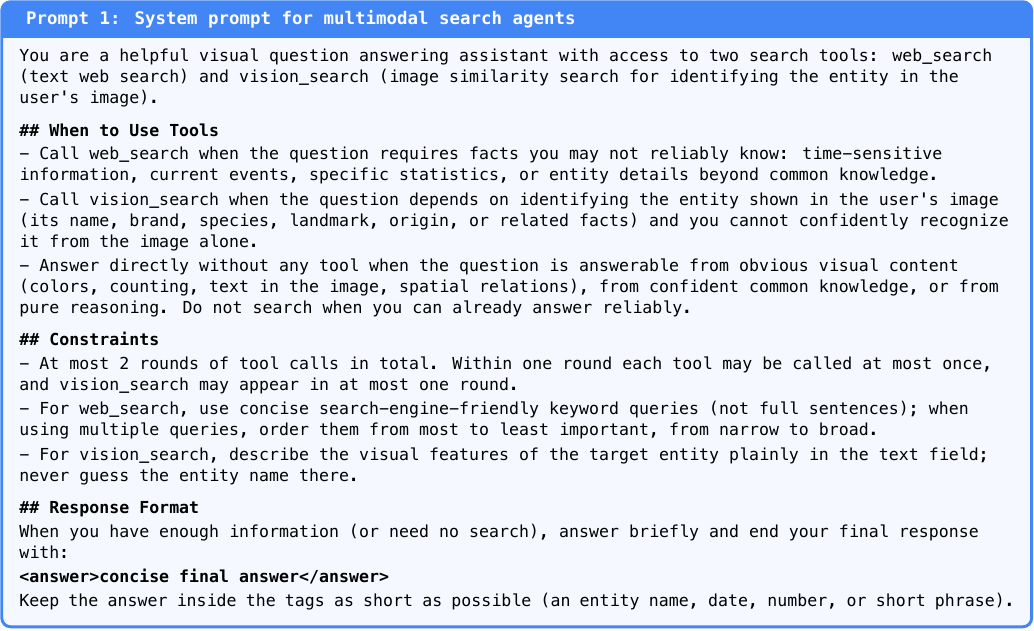}

\par\vspace{4pt}
\phantomsection\label{lst:judge-system}
\includegraphics[width=\linewidth]{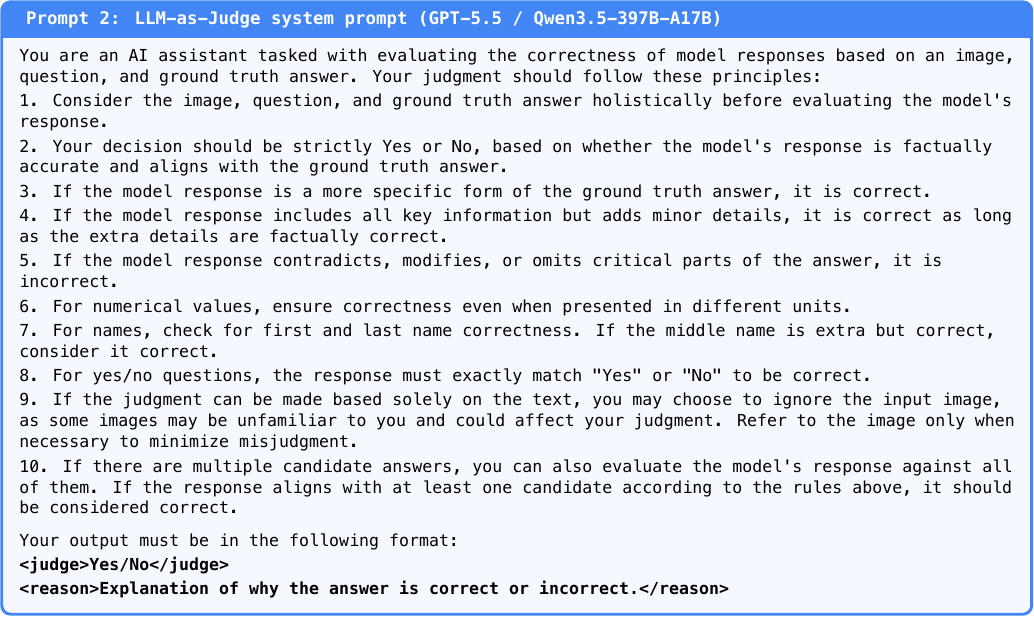}

\par\vspace{4pt}
\phantomsection\label{lst:judge-user}
\includegraphics[width=\linewidth]{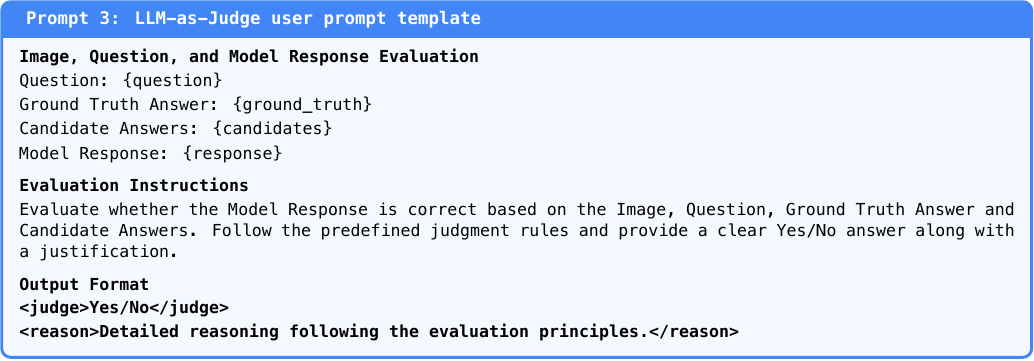}

\par\vspace{4pt}
\phantomsection\label{lst:tool-schema}
\includegraphics[width=\linewidth]{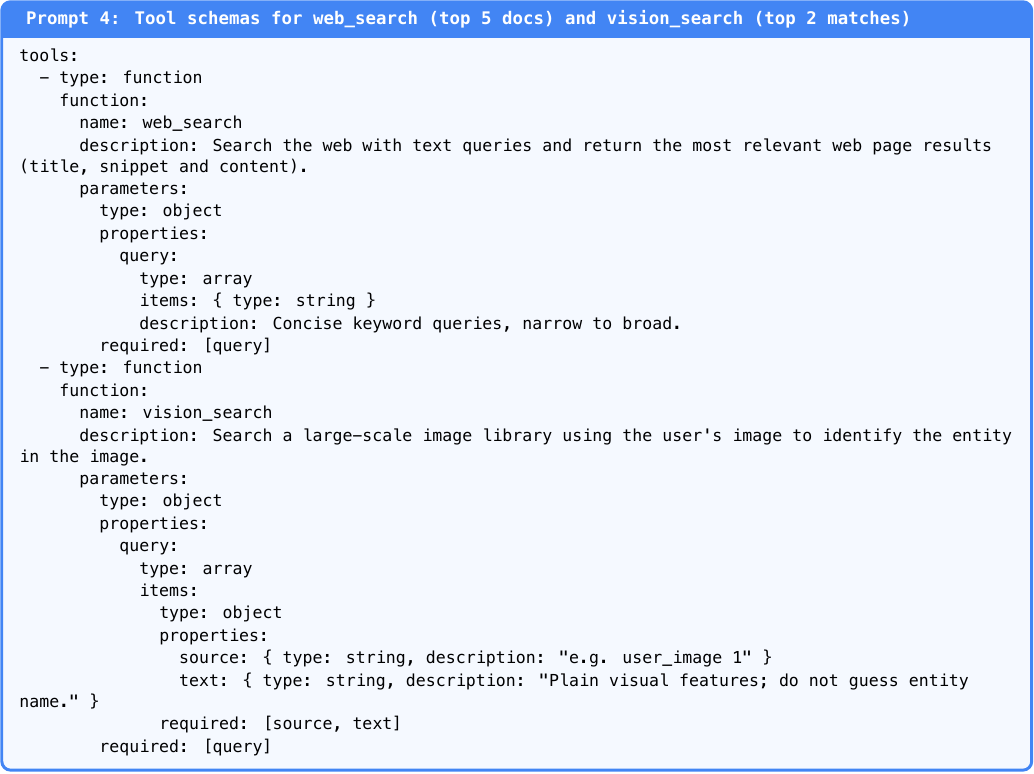}
\par\vspace{2pt}

\section{Extended Benchmark and Search-Behavior Analysis}
\label{sec:main5-audit}
This appendix reports fine-grained benchmark comparisons, tool-use breakdowns, and question-level search-state decompositions across all 4,236 evaluation examples.
Bracketed ranges $[l, u]$ report 95\% resampling ranges across 5,000 task-stratified paired question resamples on the evaluated predictions.

\subsection{Overall Paired Accuracy and Search-Rate Differences}
\label{sec:overall-paired-uncertainty}
\textbf{\method{} cuts overall search rate by roughly 20\% while matching high-accuracy RL baselines.}
Table~\ref{tab:paired-overall-ci} summarizes paired accuracy differences, 95\% resampling ranges, and search-rate reductions of \method{} relative to Vanilla GRPO and IG-Search.
Across both backbones, search rate drops by $19.52\%$--$20.16\%$, while overall accuracy differences stay within $-0.05\%$ to $+0.45\%$ and sit symmetrically inside the roughly $\pm 1.1\%$ question-sampling range around zero.

\begin{table}[!htbp]
\centering\small
\setlength{\tabcolsep}{8pt}
\caption{%
\textbf{Overall paired differences (\method{} minus comparator, \%) across all 4,236 examples.}
The 95\% Range column reports the 5,000-resample task-stratified paired range for $\Delta$Acc.%
}
\label{tab:paired-overall-ci}
\begin{tabular}{@{}llrrr@{}}
\textbf{Backbone} & \textbf{Comparator} & \textbf{$\Delta$Acc.\ (\%)} & \textbf{Acc.\ 95\% Range (\%)} & \textbf{$\Delta$SR (\%)}\\
\midrule
Qwen3.5-35B-A3B & Vanilla GRPO & $+0.40$ & $[-0.64,\; +1.44]$ & $-19.62$\\
 & IG-Search & $+0.45$ & $[-0.50,\; +1.39]$ & $-19.52$\\
\midrule
Qwen3.5-9B & Vanilla GRPO & $-0.05$ & $[-1.18,\; +1.06]$ & $-19.74$\\
 & IG-Search & $+0.33$ & $[-0.73,\; +1.42]$ & $-20.16$\\
\end{tabular}
\end{table}

\subsection{Tool-Call Breakdown and Token Usage}
\label{sec:resources}
\textbf{Skipping single-step lookups lowers total invocations and concentrates remaining calls on multi-step queries.}
Table~\ref{tab:main5-resources} breaks down total tool invocations into web and vision calls alongside conditional calls per searched example and output token counts across all 4,236 examples.
Relative to Vanilla GRPO, \method{} averages $0.23$ fewer tool calls per example on both backbones (including $0.20$--$0.22$ fewer vision searches) and fewer turns, while averaging slightly more calls on the questions where it still searches ($1.78$ vs.\ $1.66$) because multi-step queries make up a larger share of remaining searches.

\begin{table}[!htbp]
\centering\small
\setlength{\tabcolsep}{6.5pt}
\caption{%
\textbf{Per-example tool invocations and generated output tokens across all 4,236 examples.}
Calls~$|$~search divides total invocations by the number of searched examples, Tur.\ counts assistant turns including the final answer, and \textcolor{capPlotBlue}{blue} marks \method{}.%
}
\label{tab:main5-resources}
\begin{tabular}{@{}lrrrrrr@{}}
\textbf{Method} & \textbf{Calls $\downarrow$} & \textbf{Web $\downarrow$} & \textbf{Vision $\downarrow$} & \textbf{Calls $|$ search} & \textbf{Tur.\ $\downarrow$} & \textbf{Output tokens}\\
\midrule
\suppgroup{7}{Qwen3.5-35B-A3B}
SFT model & 0.947 & 0.630 & 0.317 & 1.414 & 1.95 & 170.6 \\
Direct Penalty & 0.263 & 0.076 & 0.186 & 1.117 & 1.26 & 118.1 \\
SlimSearcher & 0.194 & 0.055 & 0.139 & 1.007 & 1.19 & 17.5 \\
MMSearch-R1 & 0.701 & 0.371 & 0.331 & 1.548 & 1.70 & 138.2 \\
Vanilla GRPO & 1.581 & 0.839 & 0.742 & 1.655 & 2.58 & 224.8 \\
IG-Search & 1.542 & 0.815 & 0.726 & 1.615 & 2.54 & 225.3 \\
\rowcolor{tableOursBlue}\textbf{\method{} (Ours)} & \textbf{1.350} & 0.823 & \textbf{0.527} & 1.778 & \textbf{2.35} & 209.1 \\
\midrule
\suppgroup{7}{Qwen3.5-9B}
SFT model & 0.885 & 0.581 & 0.304 & 1.380 & 1.89 & 158.1 \\
Direct Penalty & 0.506 & 0.259 & 0.247 & 1.261 & 1.51 & 138.4 \\
SlimSearcher & 0.416 & 0.218 & 0.198 & 1.279 & 1.42 & 24.8 \\
MMSearch-R1 & 0.751 & 0.401 & 0.350 & 1.670 & 1.75 & 142.9 \\
Vanilla GRPO & 1.640 & 0.911 & 0.729 & 1.662 & 2.64 & 221.3 \\
IG-Search & 1.657 & 0.947 & 0.709 & 1.671 & 2.66 & 215.3 \\
\rowcolor{tableOursBlue}\textbf{\method{} (Ours)} & \textbf{1.411} & \textbf{0.880} & \textbf{0.531} & 1.786 & \textbf{2.41} & 217.1 \\
\end{tabular}
\end{table}

\subsection{Per-Subtask MMSearch Breakdown}
\label{sec:mmsearch-detail}
\textbf{\method{} curates search selectively across context-supplied and open-retrieval subtasks.}
Table~\ref{tab:main5-mmsearch-subtasks} breaks the pooled MMSearch column of Table~\ref{tab:main} into its four constituent subtasks.
On ReRank and Summarization, where candidate webpages or article text are already supplied in the prompt, \method{} cuts additional search rates from $92.33\%$--$99.67\%$ under Vanilla GRPO to $47.67\%$--$61.33\%$, whereas on open-ended retrieval tasks (End-to-End and MMSearch-Plus) it maintains $96.67\%$--$98.33\%$ search rates alongside competitive accuracy.

\begin{table}[!htbp]
\centering\footnotesize
\setlength{\tabcolsep}{5.5pt}
\caption{%
\textbf{Per-subtask MMSearch accuracy and search rate (\%).}
Sample-weighted pooling across all 1,211 examples yields the MMSearch column in Table~\ref{tab:main}; \textcolor{capPlotBlue}{blue} marks \method{}, and bold marks the highest Acc.\ overall and the lowest SR within the high-accuracy RL group (Vanilla GRPO, IG-Search, and \method{}).%
}
\label{tab:main5-mmsearch-subtasks}
\begin{tabular}{@{}l*{8}{r}@{}}
& \multicolumn{2}{c}{\textbf{End-to-End (300)}} & \multicolumn{2}{c}{\textbf{ReRank (300)}} & \multicolumn{2}{c}{\textbf{Summarization (300)}} & \multicolumn{2}{c}{\textbf{Plus (311)}}\\
\cmidrule(lr){2-3}\cmidrule(lr){4-5}\cmidrule(lr){6-7}\cmidrule(lr){8-9}
\textbf{Method} & Acc.\ $\uparrow$ & SR $\downarrow$ & Acc.\ $\uparrow$ & SR $\downarrow$ & Acc.\ $\uparrow$ & SR $\downarrow$ & Acc.\ $\uparrow$ & SR $\downarrow$\\
\midrule
\suppgroup{9}{Qwen3.5-35B-A3B}
SFT model & 45.00 & 87.33 & 44.00 & 32.33 & 72.33 & 35.33 & 4.18 & 54.02 \\
Direct Penalty & 35.33 & 45.33 & 41.00 & 1.00 & 72.67 & 1.67 & 6.43 & 63.34 \\
SlimSearcher & 25.33 & 32.00 & 40.00 & 0.00 & 63.00 & 0.67 & 2.89 & 33.12 \\
MMSearch-R1 & 44.67 & 81.33 & 43.67 & 14.67 & 72.00 & 12.33 & 9.00 & 82.32 \\
Vanilla GRPO & \textbf{50.67} & 100.00 & 43.67 & 92.33 & 71.00 & 99.33 & \textbf{10.93} & 100.00 \\
IG-Search & 49.67 & 99.67 & 42.33 & 93.67 & \textbf{74.67} & 98.33 & 8.36 & 100.00 \\
\rowcolor{tableOursBlue}\textbf{\method{} (Ours)} & 48.67 & \textbf{96.67} & \textbf{44.67} & \textbf{47.67} & 73.33 & \textbf{56.00} & 10.61 & \textbf{97.11} \\
\midrule
\suppgroup{9}{Qwen3.5-9B}
SFT model & 38.67 & 88.33 & 37.00 & 23.00 & 69.67 & 37.33 & 5.47 & 56.59 \\
Direct Penalty & 43.00 & 75.33 & 41.67 & 4.67 & 72.33 & 9.33 & 6.43 & 72.03 \\
SlimSearcher & 28.67 & 60.33 & \textbf{51.33} & 2.67 & 63.33 & 7.00 & 4.82 & 52.73 \\
MMSearch-R1 & 41.33 & 81.00 & 37.67 & 11.00 & 71.00 & 12.33 & 7.07 & 76.21 \\
Vanilla GRPO & 43.00 & 100.00 & 41.67 & 97.33 & 70.67 & 99.67 & 8.36 & 100.00 \\
IG-Search & \textbf{48.00} & 100.00 & 36.67 & 99.00 & \textbf{72.67} & 100.00 & 7.72 & 100.00 \\
\rowcolor{tableOursBlue}\textbf{\method{} (Ours)} & 47.33 & \textbf{98.33} & 38.33 & \textbf{50.67} & \textbf{72.67} & \textbf{61.33} & \textbf{9.97} & \textbf{98.07} \\
\end{tabular}
\end{table}

\subsection{Tasks with and without Supplied Web Context}
\label{sec:context-strata-audit}
\textbf{Over two thirds of \method{}'s omitted searches come from tasks without supplied webpage text.}
ReRank and Summarization supply webpage snippets or article text directly in the prompt ($n = 600$), whereas the remaining four tasks (SimpleVQA, DynVQA, End-to-End, and MMSearch-Plus) provide no external webpage text ($n = 3{,}636$).
As Table~\ref{tab:context-strata-audit} shows, on the 3,636 examples without supplied webpage text, \method{} lowers search rate by $15.59\%$ (35B-A3B) and $15.98\%$ (9B), accounting for $68.2\%$ and $69.5\%$ of the overall search reduction while slightly raising accuracy ($+0.19\%$ and $+0.06\%$).

\begin{table}[!htbp]
\centering\small
\setlength{\tabcolsep}{7pt}
\caption{%
\textbf{Paired differences (\method{} minus Vanilla GRPO, \%) stratified by supplied prompt web context.}
Share of $\Delta$SR gives the fraction of \method{}'s total omitted searches across all 4,236 examples contributed by each stratum.%
}
\label{tab:context-strata-audit}
\resizebox{\linewidth}{!}{%
\begin{tabular}{@{}llrrrr@{}}
\textbf{Backbone} & \textbf{Prompt web context} & \textbf{$n$} & \textbf{$\Delta$Acc.\ (\%)} & \textbf{$\Delta$SR (\%)} & \textbf{Share of $\Delta$SR (\%)}\\
\midrule
Qwen3.5-35B-A3B & Not supplied (4 tasks) & 3,636 & $+0.19$ & $-15.59$ & $68.2$\\
 & Supplied (ReRank + Summ.) & 600 & $+1.67$ & $-44.00$ & $31.8$\\
\midrule
Qwen3.5-9B & Not supplied (4 tasks) & 3,636 & $+0.06$ & $-15.98$ & $69.5$\\
 & Supplied (ReRank + Summ.) & 600 & $-0.67$ & $-42.50$ & $30.5$\\
\end{tabular}}
\end{table}

\subsection{Matched-Search-Rate Policy Mixtures}
\label{sec:matched-rate-controls}
\textbf{Input-dependent search selection outperforms every matched-rate random policy mixture.}
To test whether \method{}'s operating point can be matched by randomly mixing low- and high-search policies without input adaptation, we pair each lower-search baseline with Vanilla GRPO using an input-independent mixing weight $p$ chosen to match \method{}'s exact overall search rate ($75.92\%$ on 35B-A3B and $78.97\%$ on 9B).
Table~\ref{tab:main5-mixtures} shows that \method{} exceeds every matched-rate mixture across both backbones---including the strongest MMSearch-R1 + Vanilla GRPO mixture by $+1.00\%$ ($[+0.13,\; +1.91]$) on 35B-A3B and $+1.20\%$ ($[+0.24,\; +2.18]$) on 9B, with all eight 95\% resampling ranges strictly above zero (replacing Vanilla GRPO with IG-Search similarly yields $+1.03\%$ to $+3.82\%$ gains; separate exploratory runs without logged search rates recorded $53.28\%$ vs.\ $53.23\%$ on 9B and are excluded from matched-rate comparisons).

\begin{table}[!htbp]
\centering\small
\setlength{\tabcolsep}{6.5pt}
\caption{%
\textbf{Comparison against input-independent baseline mixtures matched to \method{}'s overall search rate (\%).}
Weight $p$ is the probability assigned to Vanilla GRPO; 95\% Range spans 5,000 paired resamples at fixed $p$.%
}
\label{tab:main5-mixtures}
\resizebox{\linewidth}{!}{%
\begin{tabular}{@{}llrrrr@{}}
\textbf{Low-search policy} & \textbf{High-search policy} & \textbf{Vanilla weight $p$} & \textbf{Mix Acc.\ (\%)} & \textbf{\method{} $-$ Mix (\%)} & \textbf{95\% Range (\%)}\\
\midrule
\suppgroup{6}{Qwen3.5-35B-A3B (\method{} Acc.\ $= 55.62\%$, matched SR $= 75.92\%$)}
MMSearch-R1 & Vanilla GRPO & 0.610 & 54.62 & $+1.00$ & $[+0.13,\; +1.91]$\\
Direct Penalty & Vanilla GRPO & 0.728 & 53.86 & $+1.76$ & $[+0.83,\; +2.67]$\\
SFT model & Vanilla GRPO & 0.313 & 53.46 & $+2.15$ & $[+1.20,\; +3.10]$\\
SlimSearcher & Vanilla GRPO & 0.743 & 51.83 & $+3.79$ & $[+2.85,\; +4.72]$\\
\midrule
\suppgroup{6}{Qwen3.5-9B (\method{} Acc.\ $= 53.23\%$, matched SR $= 78.97\%$)}
MMSearch-R1 & Vanilla GRPO & 0.633 & 52.03 & $+1.20$ & $[+0.24,\; +2.18]$\\
Direct Penalty & Vanilla GRPO & 0.663 & 51.74 & $+1.50$ & $[+0.50,\; +2.47]$\\
SFT model & Vanilla GRPO & 0.429 & 50.65 & $+2.58$ & $[+1.59,\; +3.57]$\\
SlimSearcher & Vanilla GRPO & 0.702 & 50.12 & $+3.11$ & $[+2.12,\; +4.12]$\\
\end{tabular}}
\end{table}

\subsection{Question-Level Wins and Losses on Disagreement Subsets}
\label{sec:search-disagreement}
\textbf{Retained searches drive large net wins over cost-penalized baselines, while omitted searches break even against Vanilla GRPO.}
Table~\ref{tab:main5-disagreement} examines the disagreement slices where baseline $B$ and \method{} ($O$) make opposite search decisions: state \texttt{01} ($B$ skips search while \method{} searches) and state \texttt{10} ($B$ searches while \method{} answers directly).
Against all three cost-penalized baselines on state \texttt{01} (where only \method{} searches), \method{} records far more unique correct answers than unique losses (e.g., $165$ vs.\ $105$ on 35B-A3B and $233$ vs.\ $137$ on 9B against MMSearch-R1), with every conditional and overall contribution range strictly positive.
Conversely, on the $>830$ state-\texttt{10} questions where Vanilla GRPO searches and \method{} answers directly, unique wins and losses are nearly balanced ($53$ vs.\ $49$ on 35B-A3B and $48$ vs.\ $52$ on 9B), bounding the overall accuracy impact within $[-0.57\%,\; +0.57\%]$.

\begin{table}[!htbp]
\centering\footnotesize
\setlength{\tabcolsep}{5.5pt}
\caption{%
\textbf{Question-level wins, losses, and 95\% resampling ranges (\%) on search-disagreement subsets.}
State \texttt{01} denotes \method{}-only search and state \texttt{10} denotes baseline-only search; Wins count questions answered correctly solely by that policy.%
}
\label{tab:main5-disagreement}
\resizebox{\linewidth}{!}{%
\begin{tabular}{@{}llrrrlr@{}}
\textbf{Baseline $B$} & \textbf{State $(B,O)$} & \textbf{$N$} & \textbf{\method{} Wins} & \textbf{$B$ Wins} & \textbf{Cond.\ $\Delta$Acc.\ 95\% Range (\%)} & \textbf{Overall Contrib.\ 95\% Range (\%)}\\
\midrule
\suppgroup{7}{Qwen3.5-35B-A3B}
Direct Penalty & \texttt{01} (\method{} searches) & 2,220 & 368 & 177 & $[+6.57,\; +10.65]$ & $[+3.45,\; +5.59]$ \\
SlimSearcher & \texttt{01} (\method{} searches) & 2,402 & 576 & 128 & $[+16.62,\; +20.60]$ & $[+9.40,\; +11.71]$ \\
MMSearch-R1 & \texttt{01} (\method{} searches) & 1,297 & 165 & 105 & $[+2.23,\; +7.09]$ & $[+0.68,\; +2.17]$ \\
Vanilla GRPO & \texttt{10} (\method{} skips) & 831 & 53 & 49 & $[-1.80,\; +2.84]$ & $[-0.35,\; +0.57]$ \\
IG-Search & \texttt{10} (\method{} skips) & 827 & 50 & 34 & $[-0.12,\; +4.11]$ & $[-0.02,\; +0.80]$ \\
\midrule
\suppgroup{7}{Qwen3.5-9B}
Direct Penalty & \texttt{01} (\method{} searches) & 1,645 & 291 & 156 & $[+5.74,\; +10.81]$ & $[+2.22,\; +4.20]$ \\
SlimSearcher & \texttt{01} (\method{} searches) & 1,967 & 447 & 139 & $[+13.45,\; +17.90]$ & $[+6.21,\; +8.33]$ \\
MMSearch-R1 & \texttt{01} (\method{} searches) & 1,441 & 233 & 137 & $[+4.03,\; +9.27]$ & $[+1.37,\; +3.14]$ \\
Vanilla GRPO & \texttt{10} (\method{} skips) & 836 & 48 & 52 & $[-2.80,\; +1.80]$ & $[-0.57,\; +0.35]$ \\
IG-Search & \texttt{10} (\method{} skips) & 854 & 54 & 51 & $[-2.02,\; +2.76]$ & $[-0.40,\; +0.57]$ \\
\end{tabular}}
\end{table}

\subsection{Complete Four-State Search Decomposition}
\label{sec:four-state-accounting}
For each baseline $B$ and \method{} ($O$), Table~\ref{tab:main5-four-states} partitions all 4,236 examples across the four joint search states $(B, O) \in \{0,1\}^2$.
Each state's additive contribution to the overall accuracy difference in Table~\ref{tab:main} equals its sample share ($N / 4{,}236$) multiplied by its conditional accuracy difference ($\text{Acc.\ }O - \text{Acc.\ }B$).

\begin{table}[!htbp]
\centering\footnotesize
\setlength{\tabcolsep}{6pt}
\caption{%
\textbf{Complete accuracy decomposition across joint search states $(B, O)$ over all 4,236 examples.}
Empty states ($N=0$) are omitted; for each baseline, counts sum to 4,236 and contributions sum to the overall accuracy gap in Table~\ref{tab:main}.%
}
\label{tab:main5-four-states}
\begin{tabular}{@{}llrrrrr@{}}
\textbf{Baseline $B$} & \textbf{State $(B,O)$} & \textbf{$N$} & \textbf{Share (\%)} & \textbf{Acc.\ $B$ (\%)} & \textbf{Acc.\ $O$ (\%)} & \textbf{Contribution (\%)}\\
\midrule
\suppgroup{7}{Qwen3.5-35B-A3B}
Direct Penalty & \texttt{01} & 2,220 & 52.41 & 46.89 & 55.50 & $+4.509$ \\
 & \texttt{11} & 996 & 23.51 & 31.43 & 35.04 & $+0.850$ \\
 & \texttt{00} & 1,020 & 24.08 & 75.88 & 75.98 & $+0.024$ \\
\addlinespace[2pt]
SlimSearcher & \texttt{01} & 2,402 & 56.70 & 34.47 & 53.12 & $+10.576$ \\
 & \texttt{11} & 814 & 19.22 & 29.61 & 37.47 & $+1.511$ \\
 & \texttt{00} & 1,020 & 24.08 & 69.71 & 75.98 & $+1.511$ \\
\addlinespace[2pt]
MMSearch-R1 & \texttt{01} & 1,297 & 30.62 & 57.67 & 62.30 & $+1.416$ \\
 & \texttt{11} & 1,919 & 45.30 & 39.03 & 40.28 & $+0.567$ \\
 & \texttt{00} & 1,020 & 24.08 & 76.18 & 75.98 & $-0.047$ \\
\addlinespace[2pt]
Vanilla GRPO & \texttt{10} & 831 & 19.62 & 73.41 & 73.89 & $+0.094$ \\
 & \texttt{11} & 3,216 & 75.92 & 48.69 & 49.16 & $+0.354$ \\
 & \texttt{00} & 189 & 4.46 & 86.24 & 85.19 & $-0.047$ \\
\addlinespace[2pt]
IG-Search & \texttt{10} & 827 & 19.52 & 71.95 & 73.88 & $+0.378$ \\
 & \texttt{11} & 3,216 & 75.92 & 48.88 & 49.16 & $+0.212$ \\
 & \texttt{00} & 193 & 4.56 & 88.08 & 84.97 & $-0.142$ \\
\midrule
\suppgroup{7}{Qwen3.5-9B}
Direct Penalty & \texttt{01} & 1,645 & 38.83 & 49.79 & 57.99 & $+3.187$ \\
 & \texttt{11} & 1,700 & 40.13 & 33.71 & 37.65 & $+1.582$ \\
 & \texttt{00} & 891 & 21.03 & 75.31 & 74.19 & $-0.236$ \\
\addlinespace[2pt]
SlimSearcher & \texttt{01} & 1,967 & 46.44 & 37.42 & 53.08 & $+7.271$ \\
 & \texttt{11} & 1,378 & 32.53 & 31.28 & 39.91 & $+2.809$ \\
 & \texttt{00} & 891 & 21.03 & 71.94 & 74.19 & $+0.472$ \\
\addlinespace[2pt]
MMSearch-R1 & \texttt{10} & 1 & 0.02 & 0.00 & 0.00 & $0.000$ \\
 & \texttt{01} & 1,441 & 34.02 & 51.84 & 58.50 & $+2.266$ \\
 & \texttt{11} & 1,904 & 44.95 & 36.71 & 39.44 & $+1.228$ \\
 & \texttt{00} & 890 & 21.01 & 74.94 & 74.27 & $-0.142$ \\
\addlinespace[2pt]
Vanilla GRPO & \texttt{10} & 836 & 19.74 & 73.92 & 73.44 & $-0.094$ \\
 & \texttt{11} & 3,345 & 78.97 & 47.59 & 47.65 & $+0.047$ \\
 & \texttt{00} & 55 & 1.30 & 85.45 & 85.45 & $0.000$ \\
\addlinespace[2pt]
IG-Search & \texttt{10} & 854 & 20.16 & 73.30 & 73.65 & $+0.071$ \\
 & \texttt{11} & 3,345 & 78.97 & 47.35 & 47.65 & $+0.236$ \\
 & \texttt{00} & 37 & 0.87 & 83.78 & 86.49 & $+0.024$ \\
\end{tabular}
\end{table}

\subsection{Shared-Slice Controls and Search-State Summary}
\label{sec:search-state-decomposition}
\textbf{\method{} matches Vanilla GRPO's 100\%-searched accuracy on the shared direct-answer slice.}
Section~\ref{sec:analysis} (Table~\ref{tab:search-omission-counts}) compares each method against Vanilla GRPO on its own search-omission subset; Table~\ref{tab:main5-shared-slices} instead fixes the evaluation set to the shared slice where Vanilla GRPO searches and \method{} answers directly ($N = 831$ on 35B-A3B and $N = 836$ on 9B) to compare all seven policies on identical questions.
On this shared slice, cost-penalized baselines likewise omit search almost completely ($\text{SR} \le 0.12\%$), and \method{}'s direct-answering accuracy stays within $0.48\%$ of Vanilla GRPO ($73.89\%$ vs.\ $73.41\%$ on 35B-A3B and $73.44\%$ vs.\ $73.92\%$ on 9B).
Together, Tables~\ref{tab:main5-four-states} and~\ref{tab:main5-shared-slices} show that $68\%$--$84\%$ of \method{}'s overall accuracy lead over cost-penalized baselines comes from questions where \method{} retains search while the baseline skips it (state \texttt{01}), whereas omitting search on roughly one fifth of Vanilla GRPO's searched questions incurs virtually no accuracy loss.

\begin{table}[!htbp]
\centering\small
\setlength{\tabcolsep}{7pt}
\caption{%
\textbf{All post-trained policies evaluated on the shared slice where Vanilla GRPO searches and \method{} answers directly.}
Correct counts report exact correct predictions out of $N=831$ (Qwen3.5-35B-A3B) and $N=836$ (Qwen3.5-9B); \textcolor{capPlotBlue}{blue} marks \method{}.%
}
\label{tab:main5-shared-slices}
\begin{tabular}{@{}lrrrrrr@{}}
& \multicolumn{3}{c}{\textbf{Qwen3.5-35B-A3B ($N=831$)}} & \multicolumn{3}{c}{\textbf{Qwen3.5-9B ($N=836$)}}\\
\cmidrule(lr){2-4}\cmidrule(lr){5-7}
\textbf{Method} & Correct & Acc.\ (\%) $\uparrow$ & SR (\%) $\downarrow$ & Correct & Acc.\ (\%) $\uparrow$ & SR (\%) $\downarrow$\\
\midrule
SFT model & 620 & \textbf{74.61} & 5.66 & 608 & 72.73 & 2.87 \\
Direct Penalty & 612 & 73.65 & 0.00 & 624 & \textbf{74.64} & 0.00 \\
SlimSearcher & 559 & 67.27 & 0.00 & 594 & 71.05 & 0.00 \\
MMSearch-R1 & 613 & 73.77 & 0.00 & 620 & 74.16 & 0.12 \\
Vanilla GRPO & 610 & 73.41 & 100.00 & 618 & 73.92 & 100.00 \\
IG-Search & 598 & 71.96 & 96.75 & 610 & 72.97 & 99.88 \\
\rowcolor{tableOursBlue}\textbf{\method{} (Ours)} & 614 & 73.89 & \textbf{0.00} & 614 & 73.44 & \textbf{0.00} \\
\end{tabular}
\end{table}

\section{\method{} Sensitivity and Training Controls}
\label{sec:tar-sensitivity-controls}
\textbf{\method{} maintains stable accuracy while smoothly trading off search across a fourfold $\beta$ range.}
Table~\ref{tab:tar-coefficients} reports step-50 evaluation results across penalty weights $\beta \in \{0.5, 1.0, 2.0\}$ over all 4,236 examples on both Qwen3.5-35B-A3B and Qwen3.5-9B (per-benchmark 9B bars appear in Figure~\ref{fig:beta-sensitivity}).
On Qwen3.5-9B, increasing $\beta$ lowers overall search rate monotonically from $81.42\%$ to $74.68\%$: under paired resampling relative to $\beta=1.0$, $\beta=0.5$ changes search rate by $+2.46\%$ ($95\%$ range $[+1.98\%,\; +2.95\%]$) and accuracy by $+0.35\%$ ($[-0.71\%,\; +1.42\%]$), whereas $\beta=2.0$ reduces search rate by $-4.30\%$ ($[-4.91\%,\; -3.68\%]$) with a $-0.42\%$ ($[-1.44\%,\; +0.59\%]$) accuracy change.
On Qwen3.5-35B-A3B, all three $\beta$ settings maintain overall accuracy within $55.48\%$--$55.62\%$ (every setting exceeding Vanilla GRPO's $55.22\%$) while cutting overall search rate from $95.54\%$ to $72.95\%$--$75.92\%$, with MMSearch search rate decreasing monotonically from $74.73\%$ to $73.33\%$ ($\beta=0.5$ and $2.0$ were evaluated in a standalone pass; earlier separate batch records of $55.97/78.37\%$ and $55.19/71.63\%$ on 35B and $51.11/71.40\%$ on 9B $\beta=2.0$ are kept separate across batches).

\begin{table}[!htbp]
\centering\small
\setlength{\tabcolsep}{6pt}
\caption{%
\textbf{\method{} penalty-weight sensitivity ($\beta \in \{0.5, 1.0, 2.0\}$) across all 4,236 examples (\%).}
\textcolor{capPlotBlue}{Blue} rows denote the default $\beta = 1.0$ checkpoints from Table~\ref{tab:main}; per-benchmark Qwen3.5-9B bars are shown in Figure~\ref{fig:beta-sensitivity}.%
}
\label{tab:tar-coefficients}
\resizebox{\linewidth}{!}{%
\begin{tabular}{@{}llrrrrrrrr@{}}
\textbf{Backbone} & \textbf{$\beta$} & \multicolumn{2}{c}{\textbf{SimpleVQA (2,025)}} & \multicolumn{2}{c}{\textbf{DynVQA (1,000)}} & \multicolumn{2}{c}{\textbf{MMSearch (1,211)}} & \multicolumn{2}{c}{\textbf{Overall (4,236)}}\\
\cmidrule(lr){3-4}\cmidrule(lr){5-6}\cmidrule(lr){7-8}\cmidrule(lr){9-10}
& & Acc.\ $\uparrow$ & SR $\downarrow$ & Acc.\ $\uparrow$ & SR $\downarrow$ & Acc.\ $\uparrow$ & SR $\downarrow$ & Acc.\ $\uparrow$ & SR $\downarrow$\\
\midrule
Qwen3.5-35B-A3B & 0.5 & 66.67 & 66.22 & 46.50 & 90.90 & \textbf{44.34} & 74.73 & 55.52 & 74.48\\
\rowcolor{tableOursBlue} & 1.0 & \textbf{67.51} & 68.99 & 45.60 & 91.60 & 44.01 & 74.57 & \textbf{55.62} & 75.92\\
 & 2.0 & 66.81 & \textbf{64.30} & \textbf{46.90} & \textbf{90.00} & 43.60 & \textbf{73.33} & 55.48 & \textbf{72.95}\\
\midrule
Qwen3.5-9B & 0.5 & 65.9\phantom{0} & 75.2\phantom{0} & 43.5\phantom{0} & 95.4\phantom{0} & 41.4\phantom{0} & 80.3\phantom{0} & \textbf{53.59} & 81.42\\
\rowcolor{tableOursBlue} & 1.0 & 64.59 & 72.49 & 44.10 & 94.10 & \textbf{41.78} & 77.29 & 53.23 & 78.97\\
 & 2.0 & 64.1\phantom{0} & \textbf{68.7\phantom{0}} & \textbf{44.7\phantom{0}} & \textbf{91.7\phantom{0}} & 40.6\phantom{0} & \textbf{70.6\phantom{0}} & 52.81 & \textbf{74.68}\\
\end{tabular}}
\end{table}

\textbf{Signal and scoring-schedule controls.} To examine how \method{} behaves under single-branch scoring and truncated scoring schedules, we train and evaluate two 50-step Qwen3.5-9B control variants under $\beta = 1.0$ and $\theta = 0.15$ across all 4,236 examples (Table~\ref{tab:tar-controls}).
First, replacing the paired contrast with rescaled input-only negative log-likelihood ($-\ell^{\mathrm{cf}}$, mapped to $[0,1]$ via a normal CDF with pre-fitted $\mu = 0.648, \sigma = 0.268$) yields $53.16\%$ overall accuracy and a $77.83\%$ search rate (MMSearch subtasks: End-to-End $46.33/98.00\%$, ReRank $38.67/44.33\%$, Summarization $69.33/51.00\%$, Plus $9.00/97.75\%$); as analyzed in Appendix~\ref{sec:diagnostic-protocol}, input-only likelihood captures much of the signal once empirically standardized, whereas \method{}'s paired $\ell^{\mathrm{real}} - \ell^{\mathrm{cf}}$ contrast eliminates the need to fit dataset-specific $\mu, \sigma$ parameters.
Second, an early-only schedule that runs paired scoring through update 24 and disables it from update 25 onward cuts per-step scoring time from $100$--$160$\,s to under $1$\,s (saving roughly $65$--$70$ minutes across updates 25--50) and reaches $52.55\%$ accuracy at an $80.59\%$ search rate, indicating that continued scoring in the second half of training further tightens search efficiency while retaining $+0.68\%$ accuracy (separately, a training-time permutation audit verified that shuffling penalties across eligible rollouts reassigns $7{,}400$ trajectories while preserving total penalty mass at $230.038$).

\begin{table}[!htbp]
\centering\small
\setlength{\tabcolsep}{4.6pt}
\caption{%
\textbf{Signal and scoring-schedule controls on Qwen3.5-9B across all 4,236 examples (\%).}
Both controls train for 50 updates ($\beta=1.0, \theta=0.15$) and are evaluated in standalone GPT-5.5 passes alongside the \textcolor{capPlotBlue}{blue} Table~\ref{tab:main} \method{} reference.%
}
\label{tab:tar-controls}
\resizebox{\linewidth}{!}{%
\begin{tabular}{@{}llrrrrrrrr@{}}
\textbf{Variant} & \textbf{Scoring schedule} & \multicolumn{2}{c}{\textbf{SimpleVQA (2,025)}} & \multicolumn{2}{c}{\textbf{DynVQA (1,000)}} & \multicolumn{2}{c}{\textbf{MMSearch (1,211)}} & \multicolumn{2}{c}{\textbf{Overall (4,236)}}\\
\cmidrule(lr){3-4}\cmidrule(lr){5-6}\cmidrule(lr){7-8}\cmidrule(lr){9-10}
& & Acc.\ $\uparrow$ & SR $\downarrow$ & Acc.\ $\uparrow$ & SR $\downarrow$ & Acc.\ $\uparrow$ & SR $\downarrow$ & Acc.\ $\uparrow$ & SR $\downarrow$\\
\midrule
\rowcolor{tableOursBlue}\textbf{\method{} (Ours)} & Paired contrast, steps 1--50 & 64.59 & 72.49 & 44.10 & 94.10 & \textbf{41.78} & 77.29 & \textbf{53.23} & 78.97\\
Input-only NLL & Standardized $-\ell^{\mathrm{cf}}$, steps 1--50 & \textbf{64.79} & 72.84 & \textbf{44.90} & \textbf{93.80} & 40.55 & \textbf{73.00} & 53.16 & \textbf{77.83}\\
Early-only scoring & Paired contrast, steps 1--24 & 63.95 & 75.06 & 43.30 & 95.10 & 41.12 & 77.87 & 52.55 & 80.59\\
\end{tabular}}
\end{table}

\section{Baseline Parameter Sensitivity and Training Dynamics}
\label{sec:direct-penalty-sweeps}
\textbf{Lowering cost-only penalty weights still fails to recover Vanilla GRPO or \method{} accuracy.}
Tables~\ref{tab:baseline-sensitivity-common} and~\ref{tab:baseline-sensitivity-coverage} report the per-benchmark and per-subtask results behind Figure~\ref{fig:penalty-sweeps} across all nine 50-step Qwen3.5-9B baseline configurations.
Across all tested parameter values, peak overall accuracies for Direct Penalty ($49.81\%$), SlimSearcher ($43.77\%$), and MMSearch-R1 ($50.54\%$) remain substantially below both Vanilla GRPO ($53.28\%$) and \method{} ($53.23\%$), matching Section~\ref{sec:count-comparison}'s analysis of why cost-only penalties still penalize helpful searches even at small weights.

\begin{table}[!htbp]
\centering\footnotesize
\setlength{\tabcolsep}{5pt}
\caption{%
\textbf{Baseline parameter sensitivity across benchmarks on Qwen3.5-9B at step 50 (\%).}
\textcolor{capMediumGray}{Gray} rows mark Table~\ref{tab:main} default settings, and Overall pools all 4,236 examples by sample count; evaluated in a standalone sweep pass, default MMSearch-R1 scores $50.54\%$ (vs.\ $49.88\%$ in Table~\ref{tab:main}), and every tested setting remains well below Vanilla GRPO ($53.28\%$).%
}
\label{tab:baseline-sensitivity-common}
\begin{tabular}{@{}llrrrrrrrr@{}}
\textbf{Method} & \textbf{Setting} & \multicolumn{2}{c}{\textbf{SimpleVQA (2,025)}} & \multicolumn{2}{c}{\textbf{DynVQA (1,000)}} & \multicolumn{2}{c}{\textbf{MMSearch (1,211)}} & \multicolumn{2}{c}{\textbf{Overall (4,236)}} \\
\cmidrule(lr){3-4} \cmidrule(lr){5-6} \cmidrule(lr){7-8} \cmidrule(lr){9-10}
& & Acc.\ $\uparrow$ & SR $\downarrow$ & Acc.\ $\uparrow$ & SR $\downarrow$ & Acc.\ $\uparrow$ & SR $\downarrow$ & Acc.\ $\uparrow$ & SR $\downarrow$ \\
\midrule
Direct Penalty & $c=0.01$ & 60.89 & 41.48 & \textbf{39.60} & 59.90 & 38.32 & 44.92 & 49.41 & 46.81 \\
 & $c=0.10$ & \textbf{62.12} & 39.41 & 38.70 & 56.10 & 38.40 & 40.79 & \textbf{49.81} & 43.74 \\
\rowcolor{tableModelGray} & $c=0.15$ & 59.51 & \textbf{33.28} & 36.70 & \textbf{53.40} & \textbf{40.55} & \textbf{40.63} & 48.70 & \textbf{40.13} \\
\midrule
SlimSearcher & $\lambda_C=0.3$ & \textbf{54.57} & 34.77 & 29.00 & \textbf{37.00} & \textbf{37.57} & \textbf{28.32} & 43.67 & 33.45 \\
\rowcolor{tableModelGray} & $\lambda_C=1.0$ & 53.19 & \textbf{31.06} & 28.60 & 37.50 & 36.75 & 30.88 & 42.68 & \textbf{32.53} \\
 & $\lambda_C=3.0$ & 53.78 & \textbf{30.81} & \textbf{31.20} & 47.40 & 37.41 & 35.76 & \textbf{43.77} & 36.14 \\
\midrule
MMSearch-R1 & $p=0.05$ & 62.22 & 40.15 & 39.90 & 50.40 & \textbf{39.80} & 41.37 & \textbf{50.54} & 42.92 \\
\rowcolor{tableModelGray} & $p=0.10$ & \textbf{62.67} & 40.99 & \textbf{40.10} & 52.20 & 38.89 & 45.50 & \textbf{50.54} & 44.92 \\
 & $p=0.30$ & 62.27 & \textbf{39.75} & 39.30 & \textbf{49.80} & 38.56 & \textbf{35.84} & 50.07 & \textbf{41.01} \\
\end{tabular}
\end{table}

\begin{table}[!htbp]
\centering\footnotesize
\setlength{\tabcolsep}{5pt}
\caption{%
\textbf{Per-subtask MMSearch breakdown for the Qwen3.5-9B baseline parameter sweep (\%).}
\textcolor{capMediumGray}{Gray} rows mark default settings; sample-weighted pooling across all 1,211 examples yields the MMSearch column in Table~\ref{tab:baseline-sensitivity-common}.%
}
\label{tab:baseline-sensitivity-coverage}
\begin{tabular}{@{}llrrrrrrrr@{}}
\textbf{Method} & \textbf{Setting} & \multicolumn{2}{c}{\textbf{End-to-End (300)}} & \multicolumn{2}{c}{\textbf{ReRank (300)}} & \multicolumn{2}{c}{\textbf{Summarization (300)}} & \multicolumn{2}{c}{\textbf{Plus (311)}} \\
\cmidrule(lr){3-4} \cmidrule(lr){5-6} \cmidrule(lr){7-8} \cmidrule(lr){9-10}
& & Acc.\ $\uparrow$ & SR $\downarrow$ & Acc.\ $\uparrow$ & SR $\downarrow$ & Acc.\ $\uparrow$ & SR $\downarrow$ & Acc.\ $\uparrow$ & SR $\downarrow$ \\
\midrule
Direct Penalty & $c=0.01$ & 40.67 & 82.33 & 38.67 & 6.67 & 69.33 & 12.33 & 5.79 & 77.17 \\
 & $c=0.10$ & 39.00 & 78.33 & 38.00 & 5.67 & 72.00 & \textbf{8.33} & 5.79 & \textbf{69.77} \\
\rowcolor{tableModelGray} & $c=0.15$ & \textbf{43.00} & \textbf{75.33} & \textbf{41.67} & \textbf{4.67} & \textbf{72.33} & 9.33 & \textbf{6.43} & 72.03 \\
\midrule
SlimSearcher & $\lambda_C=0.3$ & \textbf{30.00} & 61.67 & 55.00 & 3.00 & 62.00 & \textbf{4.33} & 4.50 & \textbf{43.73} \\
\rowcolor{tableModelGray} & $\lambda_C=1.0$ & 28.67 & \textbf{60.33} & 51.33 & 2.67 & \textbf{63.33} & 7.00 & 4.82 & 52.73 \\
 & $\lambda_C=3.0$ & 29.67 & 75.00 & \textbf{55.33} & \textbf{1.33} & 60.67 & 8.33 & \textbf{5.14} & 57.56 \\
\midrule
MMSearch-R1 & $p=0.05$ & 40.67 & 76.00 & \textbf{39.67} & \textbf{5.67} & \textbf{72.33} & \textbf{8.00} & 7.72 & 74.60 \\
\rowcolor{tableModelGray} & $p=0.10$ & \textbf{41.67} & 80.67 & 37.67 & 11.00 & 71.00 & 12.33 & 6.43 & 76.85 \\
 & $p=0.30$ & 40.00 & \textbf{72.67} & 37.00 & 7.00 & 69.67 & 8.33 & \textbf{8.68} & \textbf{54.66} \\
\end{tabular}
\end{table}

\begin{table}[!htbp]
\centering\small
\setlength{\tabcolsep}{8pt}
\caption{%
\textbf{Evolution of penalized fraction and mean Tool Credit Score $D$ during Qwen3.5-9B \method{} training ($\beta=1.0, \theta=0.15$).}
Penalized share divides nonzero-hinge rollouts ($D < 0.15$) by total eligible correct tool-using rollouts at that update.%
}
\label{tab:tar-dynamics}
\resizebox{\linewidth}{!}{%
\begin{tabular}{@{}rrrrr@{}}
\textbf{Update step} & \textbf{Eligible rollouts} & \textbf{Penalized rollouts ($D < 0.15$)} & \textbf{Penalized share (\%) $\downarrow$} & \textbf{Mean $D$ $\uparrow$}\\
\midrule
1 & 1,230 & 243 & 19.76 & 0.305\\
10 & 1,544 & 171 & 11.08 & 0.359\\
30 & 1,842 & 27 & 1.47 & 0.440\\
50 & 1,977 & 13 & 0.66 & 0.450\\
\end{tabular}}
\end{table}

\textbf{Within-run training dynamics.} Table~\ref{tab:tar-dynamics} tracks eligible correct search rollouts, nonzero-penalized rollouts, and mean Tool Credit Score $D$ at updates 1, 10, 30, and 50 of the Qwen3.5-9B \method{} training run.
Over the 50 updates, the penalized share among eligible correct searches drops from $19.76\%$ to $0.66\%$ while mean $D$ rises from $0.305$ to $0.450$, reflecting progressive elimination of weakly supported searches; because paired forward passes evaluate all eligible rollouts before checking $D < \theta$, per-step scoring work tracks total eligible count rather than nonzero penalty count, motivating the early-only schedule in Table~\ref{tab:tar-controls}.

\section{Case Study}
\label{sec:case-studies}

\subsection{Selective Search Omission and Multi-Step Retention}
\label{sec:case-selective}
\begin{figure}[!htb]
\centering
\includegraphics[width=\linewidth,height=0.59\textheight,keepaspectratio]{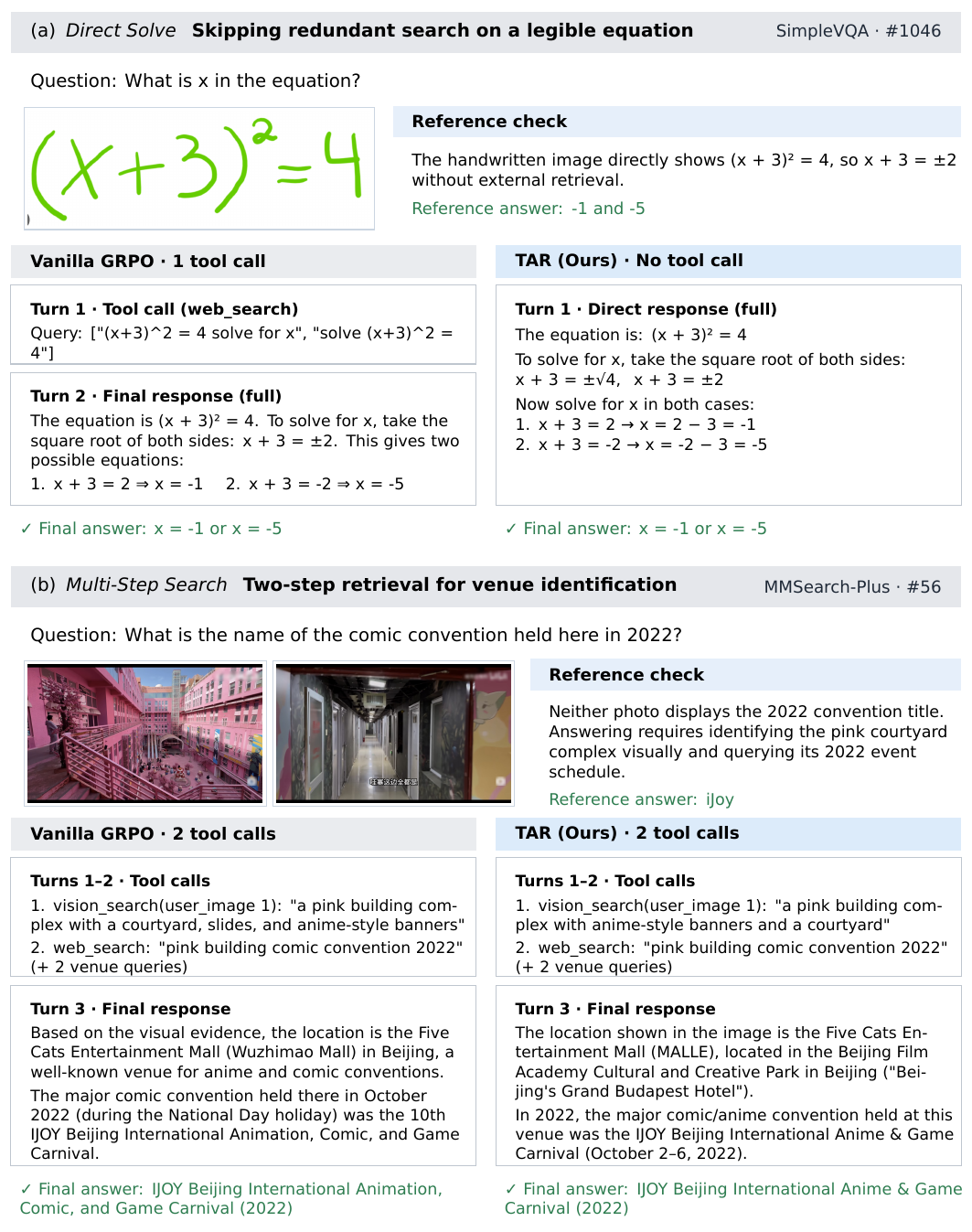}
\caption{%
\textbf{Selective search omission and multi-step retention on Qwen3.5-9B.}
In (a), where the handwritten equation is directly legible, \method{} skips Vanilla GRPO's redundant web lookup and solves the problem in a single turn; in (b), where neither venue photograph shows the 2022 convention title, \method{} retains the full two-step vision-then-web retrieval chain.
\textcolor{capCaseGreen}{Green} text marks the reference answers and verified final predictions.%
}
\label{fig:cases}
\end{figure}

Figure~\ref{fig:cases} compares paired Qwen3.5-9B evaluation trajectories from Vanilla GRPO and \method{} on two representative examples.
\method{} omits redundant web lookup when the visual equation is directly legible, while retaining two-step vision-then-web retrieval when external venue provenance is required.

\clearpage
\subsection{Failure Mode When Skipping Visual Search}
\label{sec:case-failure}
\begin{figure}[!htbp]
\centering
\includegraphics[width=\linewidth]{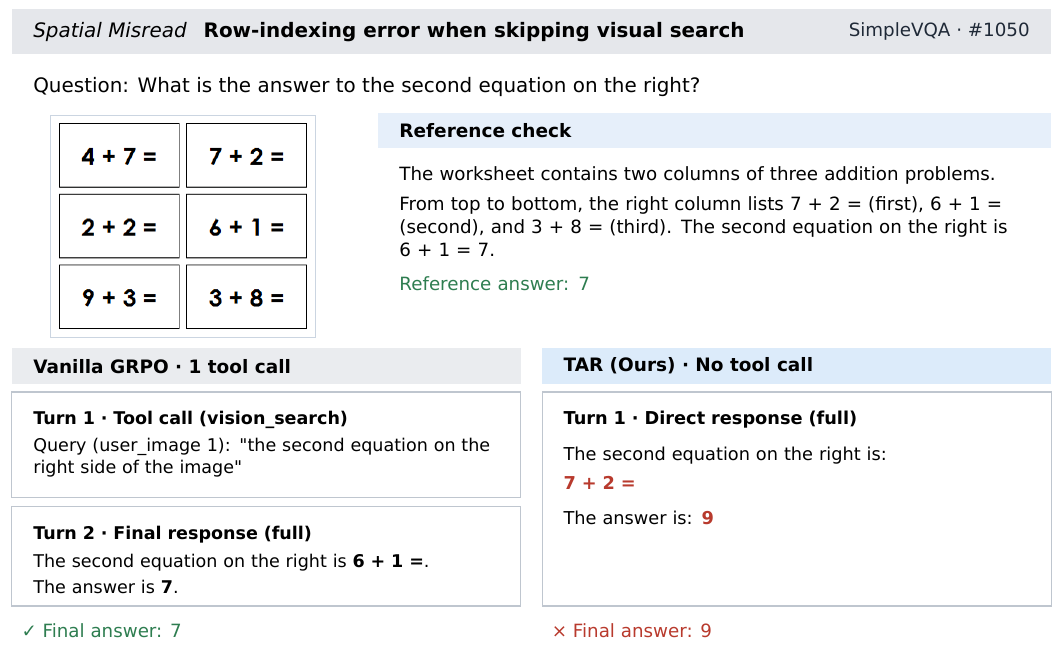}
\caption{%
\textbf{Row-indexing failure when skipping visual search on Qwen3.5-9B.}
The second equation in the right column is $6 + 1 = 7$: Vanilla GRPO invokes \texttt{vision\_search} and answers correctly, whereas \method{} answers directly from the image and misindexes the top-right equation $7 + 2 = 9$ (\textcolor{capCaseRed}{red} text marks the misread equation and resulting error).%
}
\label{fig:case-failure}
\end{figure}

Figure~\ref{fig:case-failure} shows a concrete failure case on SimpleVQA where skipping visual search leaves a row-indexing mistake uncorrected.

\section{Limitations}
\label{sec:extended-limitations}
\textbf{Methodological scope.} \method{} scores whole-trace support on completed responses via teacher forcing without decomposing per-call causal necessity across multi-step chains: masking $h_i$ removes intermediate reasoning alongside tool observations, and empirically standardized input-only likelihood ($-\ell^{\mathrm{cf}}$) attains a close accuracy--search trade-off on Qwen3.5-9B (Table~\ref{tab:tar-controls}), indicating that the two-branch formulation primarily provides parameter-free self-calibration.

\textbf{Empirical scope.} Our evaluation focuses on two Qwen3.5 backbones, three multimodal QA benchmarks, and a greedy two-round agent protocol; reported resampling ranges measure question-level variation for fixed single-seed checkpoints, leaving multi-seed RL training variance, full-panel rejudging sensitivity, and profiling of end-to-end latency, training GPU-hours, and peak memory for future work.

\end{document}